\documentclass[10pt,twocolumn,letterpaper]{article}

\usepackage[pagenumbers]{cvpr} 
\usepackage[accsupp]{axessibility}  

\usepackage{graphicx}
\usepackage{amsmath}
\usepackage{amssymb}
\usepackage{booktabs}
\usepackage{mathtools}
\usepackage{algorithm}
\usepackage{bbm}
\usepackage{dsfont}
\usepackage{algorithm,algcompatible}
\usepackage{makecell}

\DeclareMathOperator*{\argmin}{arg\,min}

\usepackage[pagebackref,breaklinks,colorlinks]{hyperref}

\usepackage[capitalize]{cleveref}
\crefname{section}{Sec.}{Secs.}
\Crefname{section}{Section}{Sections}
\Crefname{table}{Table}{Tables}
\crefname{table}{Tab.}{Tabs.}

\def\cvprPaperID{5233} 
\def\confName{CVPR}
\def\confYear{2023}

\begin{document}

\title{Marching-Primitives: Shape Abstraction from Signed Distance Function}

\author{Weixiao Liu$^{1, 2}$ \quad Yuwei Wu$^{1}$ \quad Sipu Ruan$^{1}$ \quad Gregory S. Chirikjian$^{1}$\footnotemark[1]\\
$^1$National University of Singapore \quad $^2$Johns Hopkins University\\
{\tt\small \{mpewxl, yw.wu, ruansp, mpegre\}@nus.edu.sg}
}

\maketitle

\renewcommand{\thefootnote}{\fnsymbol{footnote}}
\footnotetext[1]{Corresponding author}

\begin{abstract}
  Representing complex objects with basic geometric primitives has long been a topic in computer vision.
  Primitive-based representations have the merits of compactness and computational efficiency in higher-level tasks such as physics simulation, collision checking, and robotic manipulation.
  Unlike previous works which extract polygonal meshes from a signed distance function (SDF), in this paper, we present a novel method, named Marching-Primitives, to obtain a primitive-based abstraction directly from an SDF.
  Our method grows geometric primitives (such as superquadrics) iteratively by analyzing the connectivity of voxels while marching at different levels of signed distance.
  For each valid connected volume of interest, we march on the scope of voxels from which a primitive is able to be extracted in a probabilistic sense and simultaneously solve for the parameters of the primitive to capture the underlying local geometry.
  We evaluate the performance of our method on both synthetic and real-world datasets.
  The results show that the proposed method outperforms the state-of-the-art in terms of accuracy, and is directly generalizable among different categories and scales.
  The code is open-sourced at \url{https://github.com/ChirikjianLab/Marching-Primitives.git}.
  
\end{abstract}

\section{Introduction}
\label{sec:introduction}
Recent years have witnessed great progress in the areas of 3D shape representation and environmental perception.
Low-level representations such as surface meshes, point clouds, and occupancy grids are widely used as inputs to high-level computer vision algorithms and artificial intelligence tasks.
They have the advantage of being able to represent and visualize objects with high accuracy and rich local geometric features.
However, the low-level representations are ineffective in delivering a general and intuitive sense of structural geometry as well as part-level scene understanding.
Studies\cite{core_systems_in_human_cognition, biederman1987recognition} show that human vision, unlike computer vision, tends to perceive and understand scenes as combinations of simple primitive shapes.
Human beings perform well and robustly in complex tasks, providing a basic geometric description of the scene is available\cite{pentland1987perceptual}.
Therefore, researchers turn to exploring the possibility of interpreting complex objects and scenes with basic geometric primitives.
Taking advantage of the primitive-based representation, many higher-level tasks, such as segmentation\cite{cvxnet,Paschalidou_2021_CVPR,shapetemplates,NEURIPS2020_59a3adea}, scene understanding \cite{paschalidou2019superquadrics,paschalidou2020learning,wu2022primitive,Tulsiani_2017_CVPR}, grasping\cite{vezzani2017grasping, vezzani2018improving, quispe2015exploiting} and motion planning\cite{sipu_ral,sipu_tro}, are able to be solved efficiently.

\begin{figure} [!tp]
    \centering
    \includegraphics[width=0.95\columnwidth]{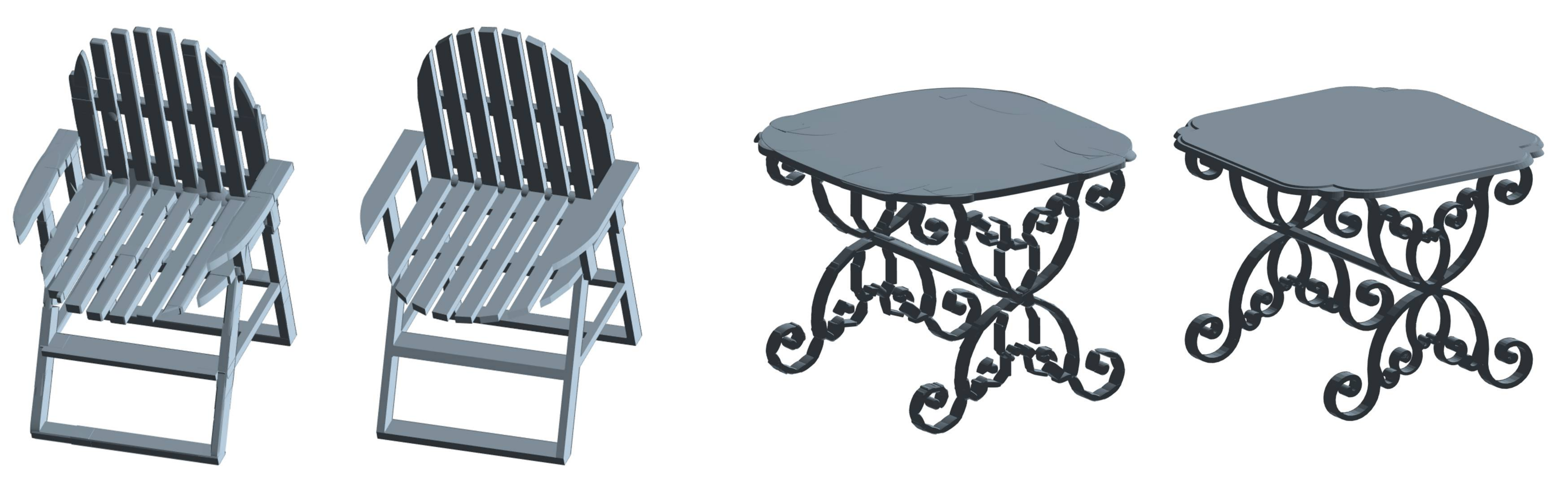} 
    \caption{Primitive-base representation versus mesh. For each pair of objects, the left one is the superquadric abstraction obtained by our algorithm, and the right one is the original mesh. The mesh of the chair is 6MB in size, while our representation only needs 4KB. An SDF representation discretized on a $128^3$ voxel grid occupies 19MB. Our abstraction is equivalent to an implicit continuous SDF, which is an approximation to the discrete SDF.}
    \label{fig:intro}
    \vspace{-0.5cm}
\end{figure}

However, it still remains challenging to extract primitive-based abstractions from low-level representations.
Starting from the 1990s, Solina \etal. \cite{Bajcsy1987ICCV, solina1990recovery, GrossICCV88} aim to extract a single superquadric representation from a simple object by minimizing the least-square error between the primitive and the measured points.
Later in \cite{leonardis1997superquadrics,  chevalier2003segmentation}, their method is extended to represent more complex objects with multiple primitives.
More recently, the authors of \cite{Liu_2022_CVPR, wu2022primitive} reformulate the task as a probabilistic inference problem with enhanced accuracy and robustness to noise and outliers.
At the same time, with the surge of data-driven techniques, researchers attempt to train neural networks to infer cuboids\cite{Tulsiani_2017_CVPR, cube_cvpr2017, cube_cvpr2018, yang2021unsupervised, smirnov2020deep} and superquadrics\cite{paschalidou2019superquadrics,paschalidou2020learning} representations in an end-to-end fashion.
However, both the computational and learning-based approaches have their own limitations.
The computational methods are vulnerable to the inherent ambiguity of the point-to-surface relationship.
For example, the algorithms tend to fill empty spaces of a non-convex object with primitives by mistake, due to the inside/outside ambiguity of a surface depicted by a set of points\cite{yang2021unsupervised,Liu_2022_CVPR}.
The main drawback of the learning approaches lies in the lack of generalizability beyond the object category on which the model is trained\cite{Liu_2022_CVPR,wu2022primitive, yang2021unsupervised,paschalidou2019superquadrics}.
Also, the shape abstraction accuracy is inferior to the computational methods.

The signed distance function (SDF) has been a successful 3D volumetric representation in varieties of computer vision and graphics tasks.
It is the basic framework for many classic 3D reconstruction algorithms such as TSDF volume reconstruction\cite{curless1996volumetric, dong2018psdf}, KinectFusion\cite{izadi2011kinectfusion}, and DynamicFusion\cite{newcombe2015dynamicfusion}.
Recently, the SDF representation is adapted to the deep learning frameworks, and exhibits boosted potentials in shape encoding\cite{park2019deepsdf,hao2020dualsdf,vasutalabot2022hybridsdf,chou2022gensdf}, surface reconstruction\cite{Weder_2020_CVPR,liao2018deep}, and shape completion\cite{rao2022patchcomplete, Dai_2017_CVPR,dai2018scancomplete}.
Usually, triangular mesh surfaces are extracted from the SDF representation with the marching cubes algorithm\cite{lorensen1987marching}.
Point cloud and occupancy grid representations are also obtained by keeping the vertices of the meshes and the sign of each voxel point, respectively.
The SDF is among the most informative 3D representations since it encodes not only the surface geometry but also the distance and side of a point relative to the shape.
Meanwhile, it is easily achievable via range images from 3D sensors\cite{curless1996volumetric}, or learnable from other input modalities\cite{park2019deepsdf, chou2022gensdf, hao2020dualsdf}.
Since we are able to extract meshes from an SDF, it is natural to think about the possibility of extracting primitives as well.
Furthermore, the primitive-based abstraction is a continuous interpretation of the complete geometric information encoded in the original discrete SDF, but requires much less storage size (Fig.\ref{fig:intro}).

Motivated by the aforementioned facts and the bottleneck of the current shape abstraction algorithms, we proposed a general shape abstraction method by reasoning directly on the informative SDF representation.
The goal of our method is to find a combination of geometric primitives whose underlying SDF values match the target values evaluated on the evenly spaced discrete grid points (Sec. \ref{sec:preliminary} and Sec. \ref{sec:primitive}).
To solve this problem, we propose a two-step iterative algorithm called the Marching-Primitives.
Our algorithm `marches' on two domains: the signed distance domain and the voxelized space domain, alternately.
Firstly, the connectivity of volumes are analyzed by generating isosurfaces on a sequence of decreasing levels of negative signed distances (Sec.\ref{sec:connectivity}).
By doing so, volumes of interest (VOIs) where primitives are likely to be encoded can be identified sequentially.
In the second step, for each of the VOIs, our algorithm marches on the neighbouring voxels to infer their probabilistic correspondences to the primitive and simultaneously optimizes the shape and pose of the primitive (Sec.\ref{sec:probabilistic_marching}).
After the primitive representation of a VOI is achieved, the fitted volumes are deactivated from the voxel grid.
Our algorithm continues marching on the signed distance domain until it approaches zero, \ie, all the interior volumes of the SDF have been captured by the recovered primitives.
We compare our algorithm with the state-of-the-art of both the computational and learning-based approaches on the ShapeNet object dataset\cite{chang2015shapenet} and D-FAUST human shape dataset\cite{dfaust:CVPR:2017} (Sec. \ref{sec:dataset}).
We also study the performance of our algorithm on different conditions(Sec. \ref{sec:resolution}).
Finally, we demonstrate the scene abstraction result of the Stanford Reading Room\cite{zhou2013elastic}, which contains several pieces of furniture of various categories(Sec. \ref{sec:scene}).


\section{Related Work}
\label{sec:related}

\textbf{SDF Representation:}
The SDF can be stored in two different ways: discrete or continuous.
A majority of computer vision and graphics algorithms are built on the SDF discretized on a 3D grid of voxel points.
The signed distances are stored on each of the corresponding voxel points.
The authors of \cite{curless1996volumetric,izadi2011kinectfusion,newcombe2015dynamicfusion} pioneer in fusing several noisy range images into a single discrete SDF.
Their work is widely applied in 3D reconstruction and plays an important role in robotics tasks such as simultaneous localization and mapping.
The discrete SDF is also a promising input/output representation for 3D deep learning \cite{rao2022patchcomplete,Dai_2017_CVPR,dai2018scancomplete}.
Recently, it becomes popular to encode shapes as a continuous SDF with neural networks\cite{park2019deepsdf}.
Vasu \etal.\cite{vasutalabot2022hybridsdf} further improve the shape encoding quality by enforcing local regularities with geometric primitives.
In \cite{chou2022gensdf}, the authors adopt a two-stage meta-learning approach to further extend the generalization capabilities of neural SDF.
With the deep neural network, it becomes possible to infer SDF representations from partially observed 3D inputs or even images.
Both the discrete and the continuous SDF are implicit representations of geometric surfaces.
To extract the explicit surface from the SDF, continuous SDFs need to be discretized on a voxel grid first and then conduct the marching cubes\cite{lorensen1987marching}.
This method allows high-quality rendering of the objects, however, surface meshes are non-sparse and contain no structural level information.
Our method provides an alternative approach to describe the underlying object in the SDF.
Instead of meshes, we directly extract a collection of sparsely parameterized primitives from the SDF.
Other than that, our primitive-based representation itself is also a concise yet continuous SDF approximation to the original discrete SDF.

\textbf{Computational Shape Abstraction:}
The most well-studied primitive for computational shape abstraction is the superquadric, due to its extensive shape vocabulary including cuboids, ellipsoids, cylinders, octohedra, and many shapes in between (\eg, cuboids with rounded edges).
It is first proposed as a versatile modeling element for complex objects in computer graphics \cite{barr1981superquadrics, pentland1987perceptual}.
Later, Solina \etal. propose a method to conduct abstraction of simple objects from range images with a single superquadric \cite{solina1990recovery, GrossICCV88}.
Leonardis \etal. \cite{leonardis1997superquadrics} and Chevalier \etal. \cite{chevalier2003segmentation} further extend the previous work to recover complex objects with multiple superquadrics with a \textit{Split-and-Merge} strategy.
A numerical instability problem is addressed and revisited in \cite{superquadrics_num_stable}.
The authors introduce an auxiliary function in the unstable region and receive a better abstraction accuracy.
More recently, Liu \etal.\cite{Liu_2022_CVPR} formulate the problem in a probabilistic fashion and propose a geometric strategy to avoid local optimum, bringing a significant improvement in robustness to outlier and fitting accuracy.
Wu \etal. \cite{wu2022primitive} extend and recast the work as a nonparametric Bayesian inference problem so as to improve the applicability on complex shapes.
To the best of the authors' knowledge, the existing computational methods are all based on range images or point clouds, which suffer from geometric ambiguities\cite{yang2021unsupervised}.
In contrast, our method takes advantage of the abundant geometric information encoded in the SDF and is easily compatible with other computer vision algorithms based on the SDF representation.

\textbf{Learning-based Shape Abstraction:}
The learning-based method is first seen in \cite{Tulsiani_2017_CVPR}.
Tulsiani \etal. propose a 3D convolutional neural network (CNN) to learn shape abstractions with cuboids from the occupancy grid.
Sun \etal.\cite{sun2019learning} design an adaptive hierarchical cuboid representation and introduce an unsupervised approach to learn to extract the parameters for the representation.
Yang \etal. \cite{yang2021unsupervised} train a variational
auto-encoder network to transform point clouds into parametric cuboids.
Other than cuboids, researchers also seek to extract spheres and ellipsoids representations from objects.
Hao \etal. \cite{hao2020dualsdf} combine the neural SDF with the spherical representation by sharing a same latent layer.
In \cite{sharma2022prifit}, ellipsoids or cuboids are extracted to help segment the input point cloud. 
However, a single type of primitive has very limited expressiveness.
Therefore, Paschalidou \etal. \cite{paschalidou2019superquadrics, paschalidou2020learning} turn to training neural networks to conduct abstractions with the superquadrics as the atomic elements.
Learning-based approaches are versatile in dealing with different input sources.
They are able to make shape abstractions from point clouds, voxel grids, or even RGB images which are so ill-conditioned that the computational approaches have little chance of working.
However, learning-based approaches rely heavily on the training dataset and thus are less generalizable to unseen categories.
Instead, our approach reasons about the primitive abstraction from a case-by-case geometric perspective, which provides an inherent advantage in generalizability and accuracy.

\section{Method}
\label{sec:method}
\subsection{Preliminary}
\label{sec:preliminary}
The discrete SDF is a volumetric surface representation built on a voxel grid $\mathbf{V}=\{\mathbf{x}_i\in \mathbb{R}^3, i=1,2,...,N\}$.
A scalar $d(\mathbf{x}_i)$ is assigned to each grid point, which indicates the signed distance of $\mathbf{x}_i$ to the nearest surface.
The point $\mathbf{x}_i$ lies inside the surface if $d(\mathbf{x}_i)<0$ and outside otherwise.
Typically, the surface mesh of an object is extracted from the SDF by the marching cubes algorithm \cite{lorensen1987marching}.
In this paper, we call the input SDF $d(\mathbf{x}_i)$ the target SDF.

For the primitive representation, we select the superquadrics\cite{barr1981superquadrics}, a family of geometric primitives defined by the implicit equation
\begin{equation}
    f(\mathbf{x})=\left(\left(\frac{x}{a_{x}}\right)^{\frac{2}{\epsilon_{2}}}+\left(\frac{ y}{a_{y}}\right)^{\frac{2}{\varepsilon_{2}}}\right)^{\frac{\varepsilon_{2}}{\varepsilon_{1}}}+\left(\frac{z}{a_{z}}\right)^{\frac{2}{\varepsilon_{1}}}=1
    \label{eqn:iofunction}
\end{equation}
The superquadric family is only encoded by 5 parameters (shape parameters $\epsilon_1,\epsilon_2\in[0, 2]\subset\mathbb{R}$, and scale parameters $a_x, a_y, a_z\in\mathbb{R}_{>0}$), but has an extensive shape vocabulary.
Note that the shape parameters can exceed 2, resulting in nonconvex shapes.
However, practically in most studies\cite{Liu_2022_CVPR,paschalidou2019superquadrics} and also in our paper, we limit them within the convex region as defined above.
The points $\mathbf{x}=[x, y, z]\in \mathbb{R}^3$ satisfying Eq. \eqref{eqn:iofunction} form the surface of the superquadric.
In this paper, we also include the Euclidean transformation $g\in SE(3)$, \ie 3 Euler angles for rotation $\mathbf{R}\in SO(3)$ and 3-dimensional translation $\mathbf{t}\in \mathbb{R}^3$, to parameterize a superquadric with a general pose.
In total, we denote a superquadric with a vector $\boldsymbol{\theta}$ of 11 elements.
According to Eq. \eqref{eqn:iofunction}, we can approximate the signed distance of a grid point $\mathbf{x}_i$ to a general posed superquadric $\boldsymbol{\theta}$ explicitly by
\begin{equation}
d_{\boldsymbol{\theta}}(\mathbf{x}_i)=\left( 1 - f^{-\frac{\epsilon_1}{2}}(g^{-1}\circ\mathbf{x}_i) \right)\|g^{-1}\circ\mathbf{x}_i\|_{2}
\label{eqn:radial_distance}
\end{equation}
We are not able to use the exact SDF of superquadrics, because it has no analytical solution and is only achievable by numerical optimization\cite{breen19983d}.
Eq.\eqref{eqn:radial_distance} is the radial distance \cite{GrossICCV88,Liu_2022_CVPR} of $\mathbf{x}_i$ to the superquadric surface, which converges to the exact signed distance as $d_{\boldsymbol{\theta}}(\mathbf{x}_i)\rightarrow0$.
Therefore, we truncate both the input target SDF $d(\cdot)$ and the primitive SDF $d_{\boldsymbol{\theta}}(\cdot)$ within the vicinity of zero to ensure the approximation accuracy.

\subsection{Problem Formulation}
\label{sec:primitive}
To obtain a primitive-based abstraction from the target SDF, we seek a combination of primitives $\boldsymbol{\Theta}\doteq\{\boldsymbol{\theta}_k, k= 1,2,...,K\}$ whose underlying signed distances measured on the voxel points (we call it the source SDF in contrast to the target SDF) match the target SDF, that is
\begin{equation}
    \boldsymbol{\Theta}=\argmin_{\boldsymbol{\Theta}}\sum_{\mathbf{x}_i\in\mathbf{V}}\min_{\boldsymbol{\theta}_k\in\boldsymbol{\Theta}} \big\|d_{\boldsymbol{\theta}_k}(\mathbf{x}_i)-d(\mathbf{x}_i)\big\|_2^2
    \label{eqn:general_goal}
\end{equation}
Our problem formulation is simple and intuitive, however, it is intractable to solve directly.
First, the number of primitives $K$ needed is unknown.
Also, the correspondences between the voxel points $\mathbf{x}_i$ and the primitives $\boldsymbol{\theta}_k$ are unknown a priori.
Therefore, it makes Eq.\eqref{eqn:general_goal} a chicken-and-egg problem:
on the one hand, if we know the set of voxel points contributing to a same primitive, we are able to solve for the parameters of the primitive by optimization;
on the other hand, if we know the configurations of the primitives, we are able to tell the belongings of each voxel.
Other than those factors, the computation complexity itself is prohibitively high, because we need to evaluate hundreds of thousands or even millions of voxel points.
This makes it infeasible to operate directly on the complete dense voxel grid.
To tackle these difficulties, we propose an iterative algorithm, the Marching-Primitives, which simultaneously solves the correspondences and primitive parameters efficiently.
It is worth noting that our method is generalizable beyond the superquadrics.
Any volumetric primitives with easily accessible SDF representations can be adapted to our framework.

\subsection{Iterative Connectivity Marching}
\label{sec:connectivity}
\begin{figure} [!tp]
    \centering
    \includegraphics[width=0.9\columnwidth]{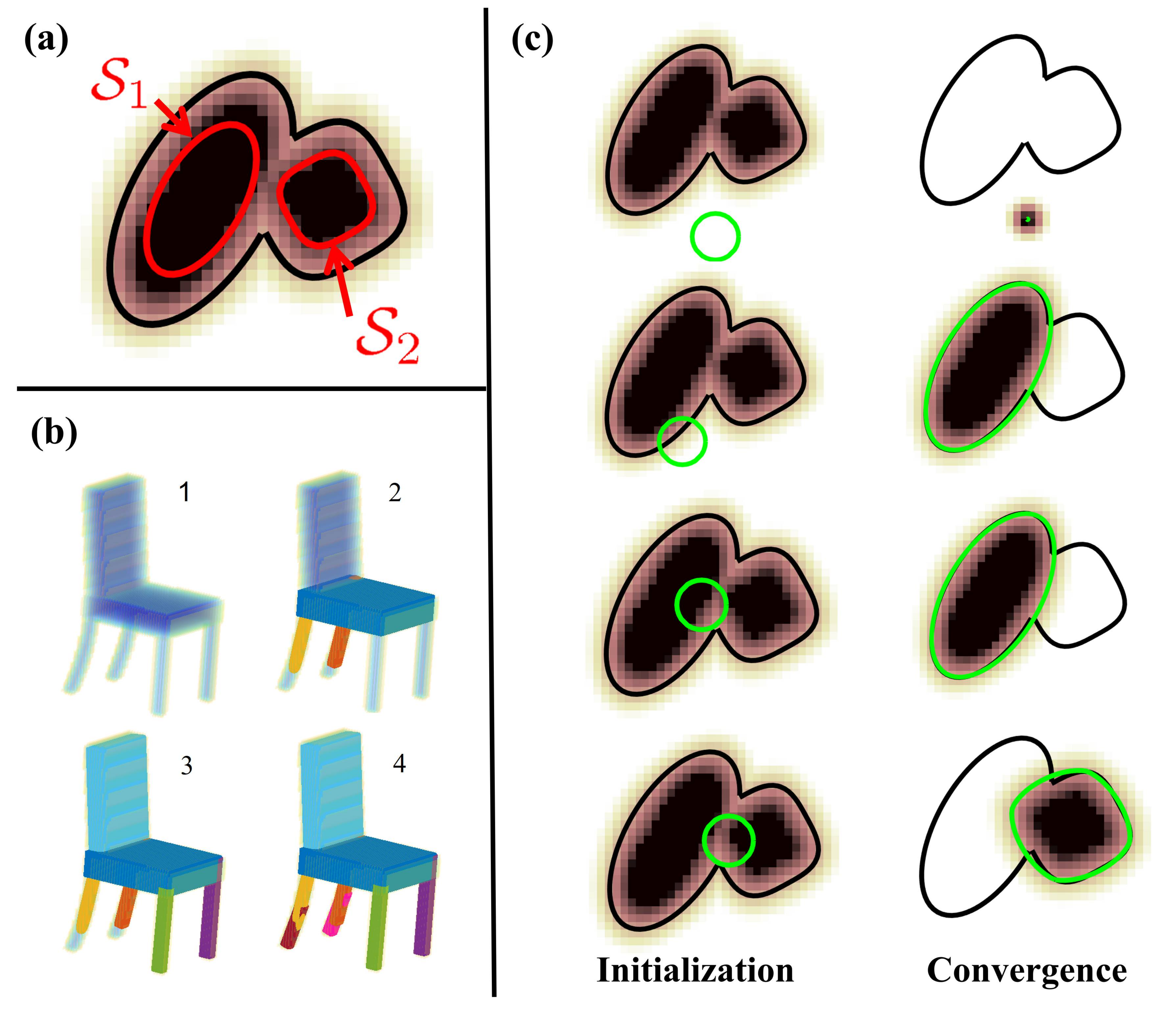} 
    \caption{Visualizations of concepts. (a) A 2D illustration of two marched isosurfaces separating the shape into two VOIs. (b) A simple example of the Marching-Primitives algorithm. The first figure is the input target SDF (voxels with signed distances larger than the truncation threshold are left blank for visualization). Figures b2-b4 illustrate the primitives marched on the sequentially detected VOIs. (c) Auto-degeneration and probabilistic marching. The green circles on the left column indicate different initialization configurations. The ones on the right are the marched results. }
    \label{fig:marching}
    \vspace{-0.5cm}
\end{figure}

Instead of setting a predefined number of primitives, our method starts from an empty set of primitives and grows as needed.
This is realized by analyzing the connectivity (in this paper, we use the 26-connectivity) of the voxels at different levels of signed distance.
Given the target SDF, our algorithm checks the connectivity of voxels whose signed distance is less than a sequence of thresholds
\begin{equation}
    T^c\doteq\{t^c_1,t^c_2,...\},\quad
    t^c_1 = \min_{\mathbf{x}_i\in \mathbf{V}} d(\mathbf{x}_i),\quad
    t^c_{m+1} = \alpha t^c_{m} 
\label{eqn:connect_init}
\end{equation}
where $\alpha\in(0,1)\subset\mathbb{R}$ is the common ratio, resulting in a geometric sequence exponentially decaying to zero.
Each threshold $t^c_m$ defines a set of disjoint isosurfaces $S_m$, encompassing the connected voxels whose signed distances are less than the threshold.
\begin{equation}
    S_m= \{\mathcal{S}_k,k = 1,2,...,{|S_m|}\}\\
\end{equation}
$|S_m|$ denotes the number of the disjoint isosurfaces.
We construct a subset $\bar{S}_m$ from $S_m$
\begin{equation}
    \bar{S}_m=\{\mathcal{S}_k\in S_m, |\mathcal{S}_k|\geq N_c\}\subseteq S_m
    \label{eqn:isosurface}
\end{equation}
where $|\mathcal{S}_k|$ is the number of the connected voxels within the isosurface $\mathcal{S}_k$.
$\bar{S}_m$ is the set of the isosurfaces encompassing no less than $N_c$ connected voxels. 
The connectivity marching starts from the innermost threshold $t^c_1$, where the resulting $\bar{S}_1$ might be empty.
The marching continues on the sequence (increasing the threshold) until $\bar{S}_m$ is non-empty.
Each of the connected volumes in $\bar{S}_m$ (we call it a VOI) is an ideal starting point for growing a primitive, because:
(1) those volumes are among the most interior of the target SDF, corresponding to most prominent part of the geometry;
(2) disconnected and weakly connected volumes can be separated apart;
(3) the SDF of a primitive can also be interpreted as layers of isosurfaces, and thus sharing similar geometric structure;
and furthermore (4) the size of the volumes is large enough to provide sufficient geometric information for the primitive recovery.
We illustrate the idea of isosurfaces in Fig.\ref{fig:marching}(a).
For each VOI, we initialize a primitive as an ellipsoid, with scales proportional to the size of its smallest bounding-box.
The details for the primitive initialization are demonstrated in the Supplementary Material.
The procedure of how a primitive marches from the initial guess to capture the local geometry is detailed in the next Sec. \ref{sec:probabilistic_marching}.
After obtaining the primitive representations for all the VOIs in $\bar{S}_m$, we subtract the voxels
\begin{equation}
    \{\mathbf{x}_i\in\mathbf{V}, d(\mathbf{x}_i)\leq 0 \land d_{\boldsymbol{\theta}}(\mathbf{x}_i)\leq0\}
\end{equation}
from the set of voxel grid $\mathbf{\mathbf{V}}$.
In other words, we deactivate the voxels which are interior of the target SDF and well fitted by a recovered primitive, while the updated SDF preserves the exterior and unfitted interior volumes.
The connectivity marching repeats with the updated target SDF, until the threshold marches higher than a preset limit close to zero, indicating that no prominent interior volume is left unrepresented.

\subsection{Probabilistic Primitive Marching}
\label{sec:probabilistic_marching}
Firstly, we explain what we mean by primitive marching.
In the marching cubes algorithm \cite{lorensen1987marching}, marching means the progressive extraction of polygonal meshes from the neighboring 8 voxel points in the discrete SDF.
In each step, the algorithm focuses within the single cube formulated by the 8 voxel points, calculates triangle vertices using linear interpolation, and then `marches' to the next one.
On the contrary, our method marches on the scope of voxels.
That is, we gradually figure out a proper collection of voxel points from which a geometric primitive is extractable.
Inspired by \cite{Liu_2022_CVPR}, we formulate this process as a maximum likelihood estimation problem with latent variables.

For each isosurface $\mathcal{S}_k \in \bar{S}_m$ obtained during the connectivity marching (Eq.\eqref{eqn:isosurface}), it is assumed that the `true' underlying primitive is encoded by $\boldsymbol{\theta}_k$, and that $z_{ik}$ is the correspondence between the primitive $\boldsymbol{\theta}_k$ and a voxel point $\mathbf{x}_i\in\mathbf{V}$.
$z_{ik}$ is a binary variable that equals 1 when the target signed distance evaluated at the $i$th voxel point results from the $k$th primitive, and 0 otherwise.
Then the target signed distance $d(\mathbf{x}_i)$ can be regarded as an observation generated from the probabilistic distribution
\begin{equation}
p(d_i|\boldsymbol{\theta}_k, z_{ik})=\Big(\frac{\mathds{1}_{d_i\in[-t,0)}}{t}\Big)^{1-z_{ik}}\mathcal{N}\big(d_i|d_{\boldsymbol{\theta}_k}(\mathbf{x}_i), \sigma^2\big)^{z_{ik}}
\label{eqn:gaussian_mixture}
\end{equation}
where $d_i$ is short for $d(\mathbf{x}_i)$; $t$ is the truncation value of the SDF; $\mathds{1}_{d_i\in[-t,0)}$ is an indicator function which equals one when $d_i\in[-t,0)$ and zero otherwise; $\mathcal{N}\big(\cdot|d_{\boldsymbol{\theta}_k}(\mathbf{x}_i), \sigma^2\big)$ is a Gaussian distribution with mean $d_{\boldsymbol{\theta}_k}(\mathbf{x}_i)$ and variance $\sigma^2$.
Consequently, our goal is to find out the optimal $\boldsymbol{\theta}_k$ and the voxel-primitive correspondences $z_{ik}$ simultaneously, which maximize the likelihood of the target SDF.

\textbf{Correspondence Marching}: We assume that $z_{ik}$ follows a Bernoulli prior distribution $B(p_0)$, independent of $\boldsymbol{\theta}_k$.
Then, given the target SDF $d_i$ and the current primitive estimation $\boldsymbol{\theta}_k$, we are able to infer the posterior correspondence $z_{ik}$ by the Bayes' rule
\begin{equation}
    p(z_{ik}|\boldsymbol{\theta}_k, d_i)=\frac{p(d_i|\boldsymbol{\theta}_k, z_{ik})p(z_{ik})}{\sum_{z_{ik}\in\{0,1\}}p(d_i|\boldsymbol{\theta}_k, z_{ik})p(z_{ik})}
\label{eqn:posterior_correspondence}    
\end{equation}
By analyzing the posterior correspondence, we can observe that our design of the probabilistic model (Eq.\eqref{eqn:gaussian_mixture}) possesses two desirable properties:
(1) voxels labeled as inside ($d_i<0$) may not contribute to the current estimation of the primitive $\boldsymbol{\theta}_k$, \ie, $p(z_{ik}=1|\boldsymbol{\theta}_k,d_i< 0)\in (0, 1)\subset \mathbb{R}$;
(2) all non-negative valued voxels always contribute to any primitives, \ie, $p(z_{ik}=1|\boldsymbol{\theta}_k,d_i\geq 0)=1$.
In other words, the primitive tries to absorb the interior voxels sharing similar local geometry, while avoids occupying any voxels labeled as the exterior.
Meanwhile, the probabilistic correspondence establishes a soft relationship between the primitive and the interior voxels, allowing part of the primitive to occupy some interior volumes without complying with the target signed distance values.
In this way, the algorithm is able to achieve an overall better fitting quality.

\textbf{Primitive Update}: Given the current estimation of the correspondences between the primitive and voxels, the parameter of the primitive is updated by
\begin{equation}
    \boldsymbol{\theta}_k=\argmin_{\boldsymbol{\theta}_k}\sum_{\mathbf{x}_i\in \mathbf{V}_{a}}P_{ik}\big\|d_{\boldsymbol{\theta}_k}(\mathbf{x_i})-d_i\big\|_2^2
\label{eqn:optimization}    
\end{equation}
where
\begin{equation}
\begin{split}
    &P_{ik} = p(z_{ik}=1|\boldsymbol{\theta}_k^{prev}, d_i)\\
    &\mathbf{V}_{a}=\big\{\mathbf{{x}_i}\in\mathbf{V}|d_{\boldsymbol{\theta}_k^{prev}}(\mathbf{x}_i)\in[-a,a]\subset \mathbb{R}\big\}
\end{split}
\label{eqn:active_set}
\end{equation}
$\boldsymbol{\theta}_k^{prev}$ is the previous estimation of the primitive parameters.
$\mathbf{V}_a$ is an adaptive subset of the complete voxels $\mathbf{V}$, which includes the voxels close to the surface of the previous estimated primitive.
$a$ defines the distance threshold of the activated region.
Only voxels in $\mathbf{V}_a$ are activated during the optimization.
The reason why we use an adaptive subset instead of the complete voxel space is twofold.
As discussed in Sec. \ref{sec:preliminary}, both the source and the target SDF are truncated within a vicinity of zero.
Therefore, small variations around $\boldsymbol{\theta}_k^{prev}$ (\ie, small changes on the shape of the primitive surface) have minor if not zero effects on the source SDF values evaluated at voxels distant to the primitive surface.
Moreover, the size of $\mathbf{V}_a$ is much smaller than the complete set, and thus providing a significant boost in performance.

The correspondence marching and the primitive updating alternate until convergence.
This process is akin to the EM algorithm \cite{DEMP1977}.
The scope of voxels from which a primitive can be extracted is `marched' with the progressive variation of the primitive shape, and simultaneously the parameters of the primitive are optimized based on the scope of voxels.
We apply the optimization method proposed in \cite{Liu_2022_CVPR} to avoid local minima.
More derivation and implementation details can be found in the Supplementary Material.

\subsection{Fail-safe Auto-degeneration}
\label{sec:primitive_degeneration}
The primitives are roughly initialized in the connectivity marching step and further grow to capture local geometric shapes.
Since the probabilistic marching is based on optimization, an important question to ask is if the recovered primitives can always converge to a proper shape, regardless of poor initialization.
We demonstrate that our probabilistic marching strategy is robust to initialization.
Firstly, our method possesses a feature called auto-degeneration- \ie the primitive will autonomously degenerate towards a point when accidentally initialized far from the target volumes.
This is because, under this circumstance, only variations which shrink the volume of the primitive can decrease the difference between the truncated target and source SDF.
Secondly, when the primitive is initialized near or inside the target volumes, the marching process encourages the primitive to converge to one nearest local target shape by analyzing the posterior correspondences.
Fig.\ref{fig:marching}(c) illustrates the examples of the above properties.
After the primitive marching, our algorithm removes the degenerated primitives.
As a fail-safe measure, primitives which significantly contradict the target SDF is also removed.
Detailed implementations can be found in the Supplementary Material.

\begin{figure*}[!th]
    \centering 
    \includegraphics[scale=0.055]{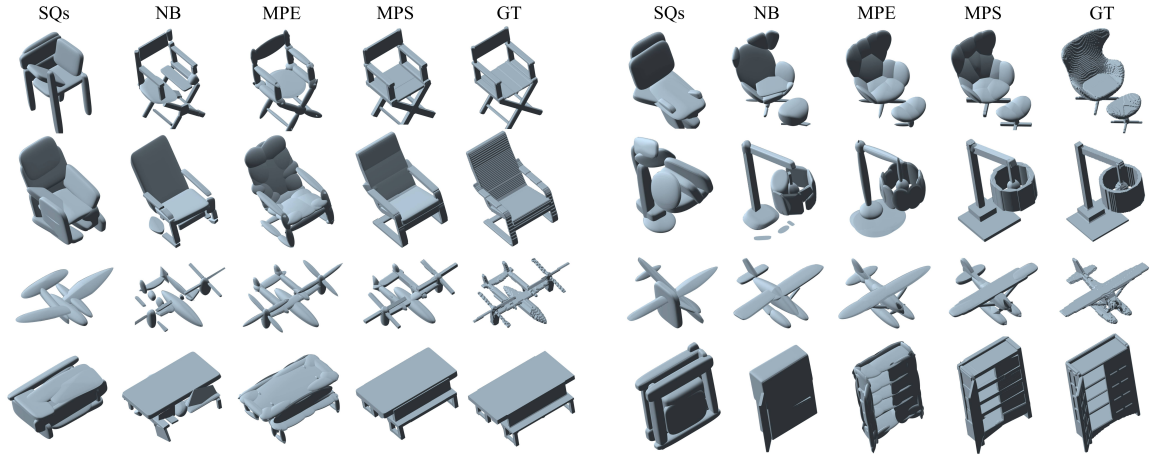} 
    \caption{Shape Abstraction results on the ShapeNet dataset. From left to right: SQs\cite{paschalidou2019superquadrics}, Non-parametric Bayesian (NB)\cite{wu2022primitive}, Marching-Primitives with ellipsoids (MPE), Marching-Primitives with superquadrics (MPS), and the ground truth (pre-processed watertight mesh).}
    \label{fig:shapenet}
\vspace{-0.4cm}
\end{figure*}
\section{Experiments}
\label{sec:experiments}
In this section, we conduct several experiments to demonstrate the high accuracy and generalizability of our proposed method.
First, we show the shape abstraction results on various datasets, ranging from daily objects to human body models.
Furthermore, we study the impact of the grid resolution and the truncation threshold on our algorithm.
In the end, we apply our algorithm on a large-scale complex indoor scene, which contains a mixed combination of furniture such as sofas, tables, books, lamps, etc.
Implementation details can be found in the Supplementary.

\subsection{Evaluation on Datasets}
\label{sec:dataset}
\textit{Baselines}: We compare our method with both the state-of-the-art learning-based method\cite{paschalidou2019superquadrics} and the computational method \cite{wu2022primitive}, which infer superquadric abstractions of the input objects.
For convenience, we refer to \cite{paschalidou2019superquadrics} as SQs, and \cite{wu2022primitive} as NB.
We use the official codes and follow the implementation details as stated in these papers, respectively.
We do not compare with the cuboid-based algorithms\cite{cube_cvpr2017,sun2019learning,yang2021unsupervised}, since they focus more on the semantic level abstraction and are limited in expressiveness and accuracy.
A detailed performance comparison between the cuboid and superquadric abstractions can be found in \cite{paschalidou2019superquadrics}.

\textit{Datasets}: We evaluate on two commonly used datasets, the ShapeNet\cite{chang2015shapenet} and the DFAUST\cite{dfaust:CVPR:2017}.
In both of the datasets, we are only provided with triangular meshes.
Therefore, we need to transform the meshes into discrete SDF representations.
It is straightforward to calculate the signed distance value of a point to a watertight mesh.
However, the ShapeNet is a human-made synthetic dataset containing many non-watertight meshes formed by non-volumetric 2D surfaces, self-intersecting or overlapped triangles.
Therefore, we pre-process the original ShapeNet models into watertight meshes and then generate the SDF representation with the fast marching algorithm, following the same procedure in \cite{chou2022gensdf}.
In consideration of fairness and consistency, we use the pre-processed watertight meshes as the common ground truth and the inputs to SQs and NB.
For the DFAUST dataset, we generate the SDF directly since the provided meshes are already watertight.

\textit{Metrics}: Following \cite{wu2022primitive,paschalidou2019superquadrics} we use the Chamfer-L1 distance and the volumetric intersection over union (IoU) as the quantitative evaluation metrics.
The computation of the metrics is detailed in the Supplementary Material.

\begin{table*}[!ht]
    \centering
    \begin{tabular}{c||cccc||cccc} 
    \Xhline{1.0pt} 
    & \multicolumn{4}{c||}{Chamfer-$L_{1}$} & \multicolumn{4}{c}{IoU}\\
    Category & SQs\cite{paschalidou2019superquadrics} & NB\cite{wu2022primitive} & MPE(Ours) & MPS(Ours) & SQs\cite{paschalidou2019superquadrics} & NB\cite{wu2022primitive} & MPE(Ours) & MPS(Ours)\\
    \hline
    airplane   & 0.037 & 0.023 & 0.021 & $\boldsymbol{0.019}$ &  0.441 & 0.671 & 0.731 & $\boldsymbol{0.768}$\\
    bench      & 0.056 & 0.028 & 0.020 & $\boldsymbol{0.020}$ &  0.238 & 0.579 & 0.730 & $\boldsymbol{0.819}$\\
    bottle     & 0.047 & 0.033 & 0.026 & $\boldsymbol{0.017}$ &  0.686 & 0.665 & 0.886 & $\boldsymbol{0.924}$\\
    cabinet    & 0.059 & 0.036 & 0.037 & $\boldsymbol{0.028}$ &  0.394 & 0.666 & 0.840 & $\boldsymbol{0.948}$\\
    can        & 0.066 & 0.036 & 0.036 & $\boldsymbol{0.022}$ &  0.706 & 0.553 & 0.908 & $\boldsymbol{0.950}$\\
    chair      & 0.068 & 0.027 & 0.023 & $\boldsymbol{0.020}$ &  0.300 & 0.685 & 0.785 & $\boldsymbol{0.871}$\\
    lamp       & 0.072 & 0.029 & 0.022 & $\boldsymbol{0.021}$ &  0.234 & 0.589 & 0.750 & $\boldsymbol{0.802}$\\
    speaker    & 0.064 & 0.041 & 0.037 & $\boldsymbol{0.033}$ &  0.346 & 0.656 & 0.858 & $\boldsymbol{0.920}$\\
    mailbox    & 0.095 & 0.026 & 0.026 & $\boldsymbol{0.024}$ &  0.333 & 0.694 & 0.802 & $\boldsymbol{0.905}$\\
    rifle      & 0.038 & 0.020 & 0.019 & $\boldsymbol{0.019}$ &  0.446 & 0.732 & 0.744 & $\boldsymbol{0.811}$\\
    sofa       & 0.054 & 0.037 & 0.029 & $\boldsymbol{0.023}$ &  0.497 & 0.726 & 0.857 & $\boldsymbol{0.940}$ \\
    table      & 0.070 & 0.024 & 0.024 & $\boldsymbol{0.022}$ &  0.247 & 0.745 & 0.818 & $\boldsymbol{0.932}$\\
    phone      & 0.040 & 0.021 & 0.023 & $\boldsymbol{0.021}$ &  0.681 & 0.872 & 0.891 & $\boldsymbol{0.947}$\\
    watercraft & 0.048 & 0.032 & 0.022 & $\boldsymbol{0.022}$ &  0.465 & 0.618 & 0.793 & $\boldsymbol{0.836}$\\
    \hline 
    mean       & 0.057 & 0.028 & 0.024 & $\boldsymbol{0.022}$ &  0.368 & 0.674 & 0.793 & $\boldsymbol{0.870}$\\
    \Xhline{1.0pt} 
    \end{tabular}
    \caption{Quantitative results on Shapenet. MPE and MPS are short for Marching-Primitives with ellipsoids and superquadrics, respectively}
    \vspace{-0.6cm}
    \label{table:shapenet}
\end{table*}

\textbf{Results on ShapeNet}: 
We experiment on 14 categories from the ShapeNet dataset.
We split the dataset randomly into the training (80\%) and testing (20\%) sets\cite{choy}, where we train one SQs model per category on the training sets and evaluate all the methods on the testing sets.
For our method, we use the SDF representation discretized on a voxel grid of size $100^3$ and range $[-0.5, 0.5]^3$.
Other than superquadrics, we also test our method with ellipsoids as the base primitive, which is a special case of the superquadric representation when the shape parameters $\epsilon_1, \epsilon_2$ are fixed to 1.
The quantitative results are summarized in Table. \ref{table:shapenet}.
We also demonstrate the qualitative comparison among different shape abstraction methods in Fig.\ref{fig:shapenet}.
Our method outperforms all the baselines, even implemented with less expressive ellipsoids.
The computational method generally performs better than the learning-based method.
This is because the parameters of the primitive are so sparse and geometrically interrelated that they are difficult to get mapped from a high dimensional input by a neural network.
Compared with NB, our method has richer details, clearer edges, and more importantly does not occupy the exterior space. 
Two factors contribute to the advantages.
The first one is the extra geometric information embedded in the SDF representation, which eliminates the inherent interior/exterior ambiguity of point clouds.
The other one is our special design of the probabilistic generative model, as discussed in Sec.\ref{sec:probabilistic_marching}.
It is interesting to observe that our algorithm can extract abstractions from objects not intuitively depictable by the primitives.

\begin{figure} [!tp]
    \centering
    \includegraphics[width=0.9\columnwidth]{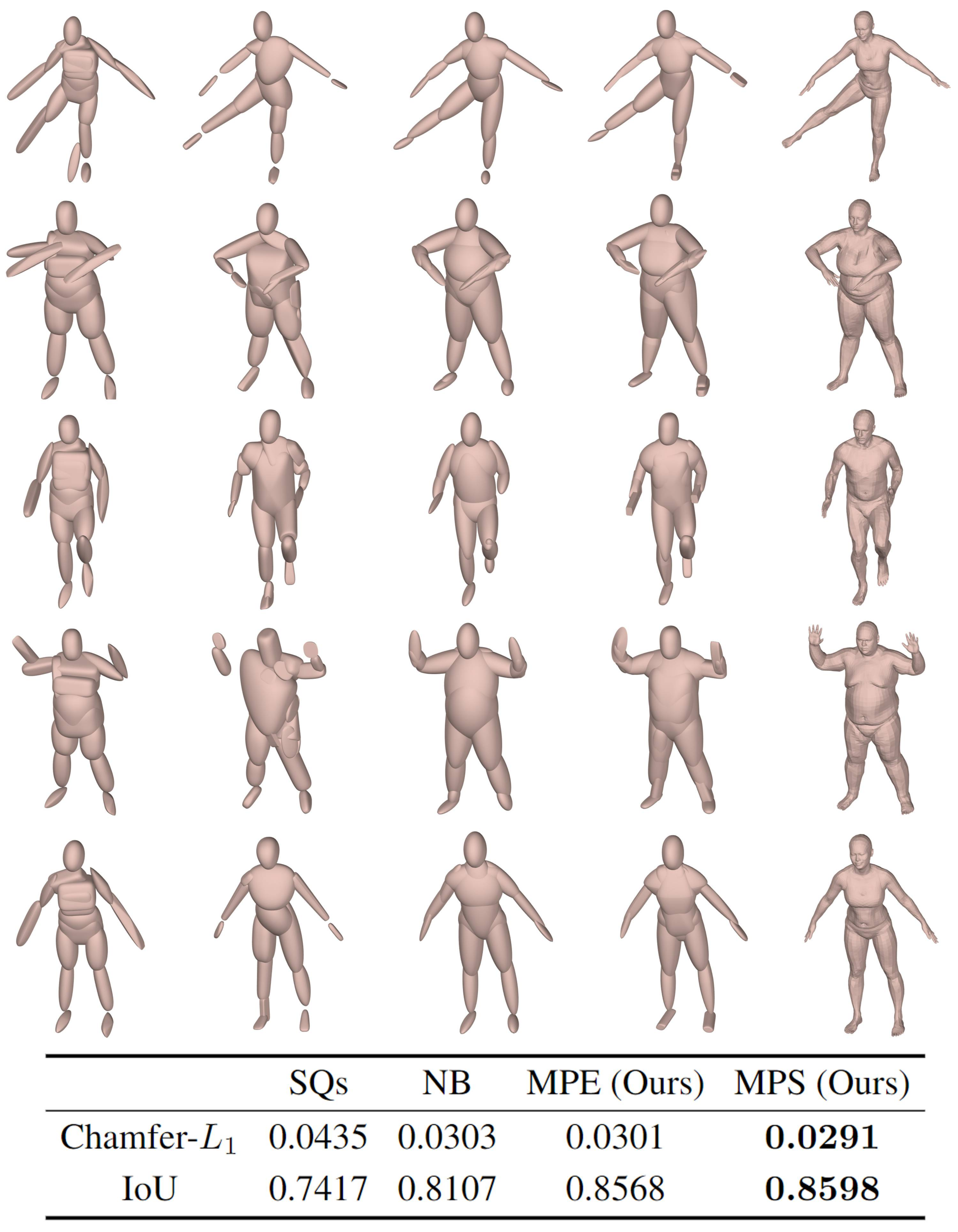} 
    \caption{Shape abstraction results on the DFAUST dataset. From left to right: SQs\cite{paschalidou2019superquadrics}, NB\cite{wu2022primitive}, MPE, MPS, and the ground truth.}
    \label{fig:dfaust}
    \vspace{-0.5cm}
\end{figure}

\textbf{Results on DFAUST}:
We also evaluate on the DFAUST dataset of human body models.
We follow the split settings in\cite{paschalidou2020learning} to train the learning-based method and evaluate all the methods on the testing set.
The SDF is discretized on a grid of size $64^3$ and side length $1$.
Similar to the ShapeNet experiment, we test our method with both superquadrics and ellipsoids.
The results are shown in Fig.\ref{fig:dfaust}.
Our method can accurately capture various postures, while the baselines fail to distinguish different body parts in some cases.
The abstraction quality of SQs is much better on this dataset compared with the ShapeNet, since human bodies share a common articulated structure that can be captured by the neural network.
We observe that the ellipsoid abstraction also achieves satisfying accuracy.
This is because the human body mostly consists of rounded shapes compared with man-made objects in the ShapeNet.

\begin{figure*} [!tp]
    \centering
    \includegraphics[scale=0.078]{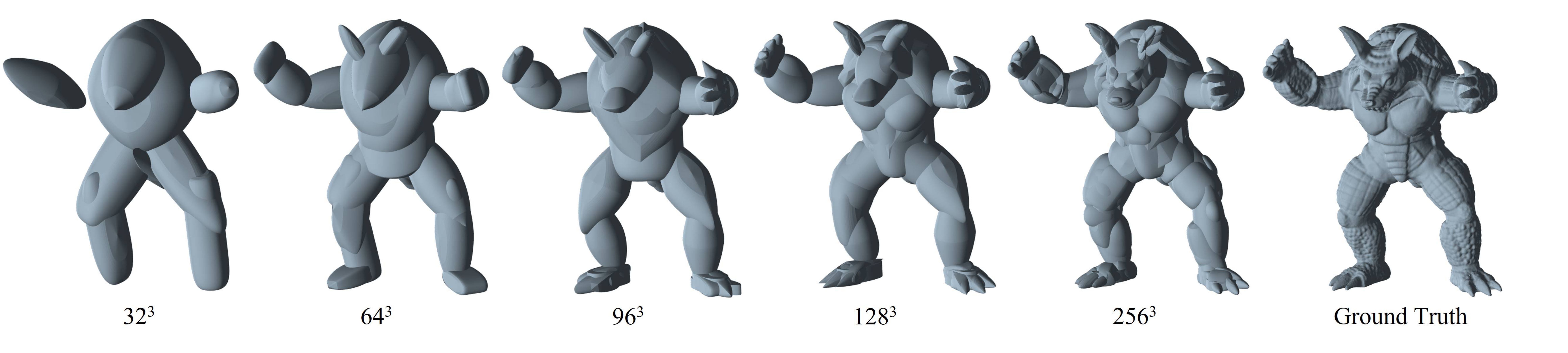} 
    \caption{Visualization of the shape abstractions on different grid resolutions. Abstraction accuracy increases with the input grid resolution.}
    \label{fig:armadillo}
    \vspace{-0.6cm}
\end{figure*}

\subsection{Performance Study}
\label{sec:resolution}
\begin{figure}[!ht]
    \centering 
    \includegraphics[width=0.97\columnwidth]{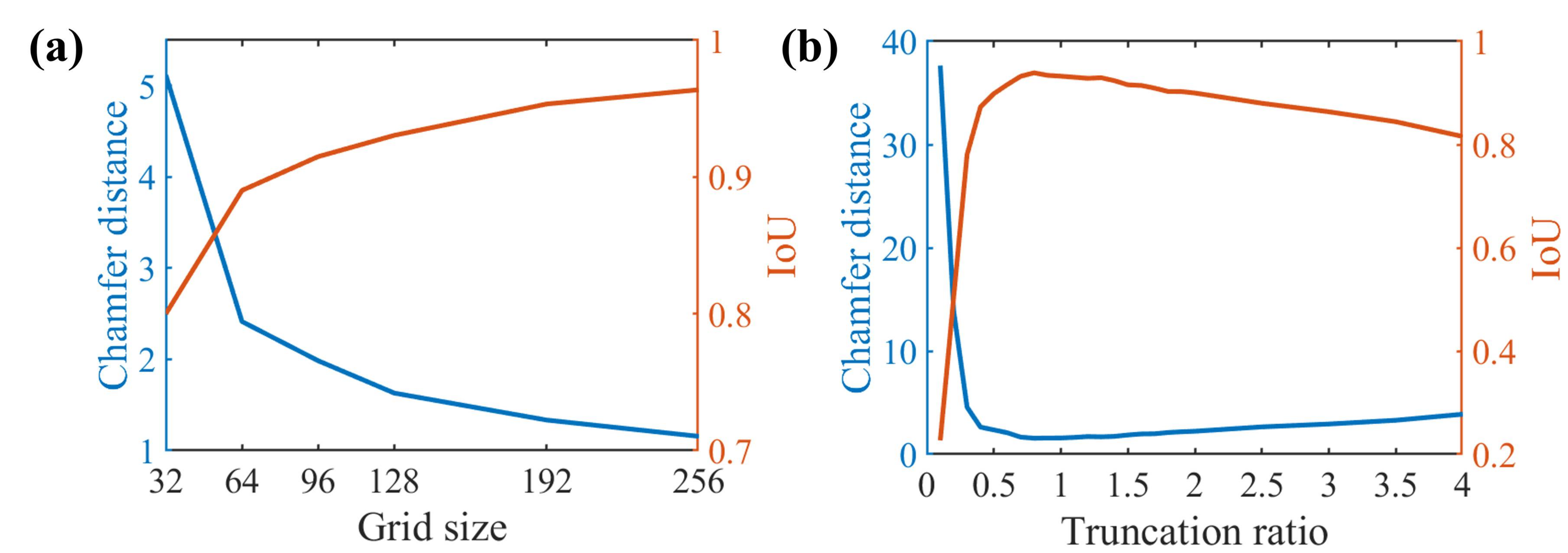} %
    \caption{Quantitative results on shape abstraction accuracy. (a) Abstraction accuracy versus input grid resolution. (b) Abstraction accuracy versus truncation ratio, evaluated at grid resolution $128^3$. We evaluate on two metrics: Chamfer-$L_{1}$ and IoU.}
    \label{fig:armadillo_curve}
\vspace{-0.4cm}
\end{figure}

This section investigates how the Marching-Primitives algorithm behaves with different voxel grid sizes and distance truncation thresholds.
We experiment on the Armadillo from the Stanford 3D scanning repository\cite{curless1996volumetric}.
First, we discretize the model on the voxel grids of different resolutions.
The qualitative abstraction results are visualized in Fig. \ref{fig:armadillo}, and the quantitative results are summarized in Fig. \ref{fig:armadillo_curve}(a).
When the input grid resolution is low, the recovered primitive representation is relatively abstract.
With the increase of the grid resolution, our method is able to extract detailed abstraction more faithful to the target shape.
This is because the higher the resolution, the more geometric information our method can utilize to guide the primitive marching process.
As discussed in Sec.\ref{sec:preliminary}, we use a truncated version of the target and source SDF.
Therefore, we also evaluate the performance on different truncation thresholds, starting from 0.1 to 4 times the interval of the voxel grids.
The results are shown in Fig.\ref{fig:armadillo_curve}(b).
The abstraction accuracy increases as the threshold decreases at first.
The reason lies in that the approximated superquadric (source) SDF converges to the true value when truncated close to the surface.
If we further decrease the threshold, however, we observe an acute decrease in accuracy.
This is because an overly small truncation threshold corrupts the original geometric information embedded in the target SDF, as well as the smoothness of the cost function (Eq. \eqref{eqn:optimization}).

\subsection{Scene Abstraction}
\label{sec:scene}
\begin{figure}[!ht]
    \centering 
    \includegraphics[width=0.73\columnwidth]{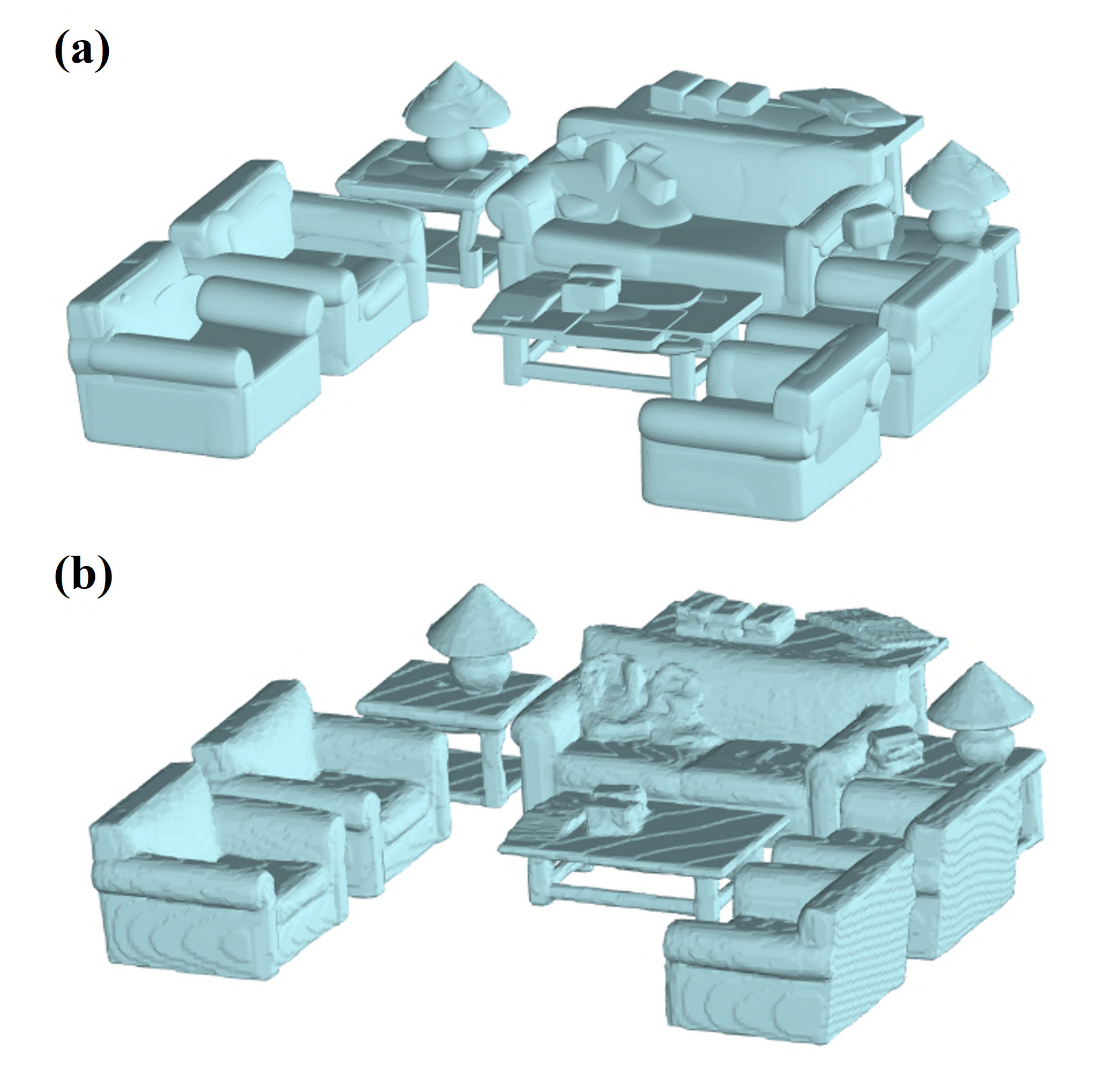} 
    \caption{Qualitative result of the scene abstraction. (a) Superquadric abstraction by the Marching-Primitives. (b) Mesh generated by the marching cubes\cite{lorensen1987marching} from the same SDF.}
    \label{fig:readingroom}
\vspace{-0.4cm}
\end{figure}

Other than single objects, we also test the capability of our method in representing complex scenes with geometric primitives.
We experiment on a real-world indoor scene called \textit{Reading Room}\cite{zhou2013elastic}.
The scene is captured by an Asus Xtion Pro Live camera.
The RGB-D scans are fused utilizing\cite{izadi2011kinectfusion} (an SDF-based 3D reconstruction algorithm), and further fine-tuned with the method proposed in \cite{zhou2013elastic}.
However, the SDF representation of the scene is not available publicly.
Therefore, we pre-process the scene mesh by removing the floor plane, filling the holes, and transforming it into the SDF representation with grid size $400^3$.
The scene abstraction task is much more difficult compared with the object-level task because various items with great differences in size and shape are present in the very same space.
We compare our abstraction result with the mesh extracted by the marching cubes algorithm on the same discrete SDF representation, as shown in Fig.\ref{fig:readingroom}.
Our representation is not only visually satisfying but also contains abundant geometric information the mesh lacks, since it is a continuous SDF approximation to the input discrete SDF.
Moreover, our highly sparse representation is only 8.2KB in size, while the discrete SDF occupies 203MB.

\section{Conclusion}
\label{sec:conclusion}
We propose the Marching-Primitives, the first algorithm to extract primitive-based abstractions from the volumetric SDF representation.
Our method outperforms the state-of-the-art in terms of accuracy and generalizability on both synthetic and real-world datasets.
Our primitive-based representation is sparse, accurate, generalizable, and can be expressed analytically without training, which we believe will facilitate and inspire further explorations in scene understanding, 3D reconstruction, and robot motion planning.
However, our algorithm has the limitation of not being able to properly extract object parts thinner than the interval of the voxel grid.
This problem can be solved by either increasing the grid resolution or detecting and thickening the invalid parts.
Future work includes the parallelization of the marching process and inferring semantic-level interpretations from the current primitive-based geometric features. 

\vspace{\baselineskip}
\noindent \textbf{Acknowledgements}
This research is supported by NUS Startup  grants A-0009059-02-00 and A-0009059-03-00, National Research Foundation, Singapore, under its Medium Sized Centre Programme - Centre for Advanced Robotics Technology Innovation (CARTIN), subaward A-0009428-08-00, and AME Programmatic Fund Project MARIO A-0008449-01-00, and JHU internal funds.

{\small
\bibliographystyle{ieee_fullname}
\bibliography{egbib}

\begin{thebibliography}{10}\itemsep=-1pt

\bibitem{Bajcsy1987ICCV}
R. Bajcsy and F. Solina.
\newblock Three dimensional object representation revisited.
\newblock In {\em Proceedings of the IEEE/CVF International Conference on
  Computer Vision (ICCV)}, pages 231--240, 1987.

\bibitem{barr1981superquadrics}
A.~H. Barr.
\newblock Superquadrics and angle-preserving transformations.
\newblock {\em IEEE Computer graphics and Applications}, 1(1):11--23, 1981.

\bibitem{biederman1987recognition}
I. Biederman.
\newblock Recognition-by-components: a theory of human image understanding.
\newblock {\em Psychological review}, 94(2):115, 1987.

\bibitem{dfaust:CVPR:2017}
F. Bogo, J. Romero, G. Pons-Moll, and M.~J. Black.
\newblock Dynamic {FAUST}: {R}egistering human bodies in motion.
\newblock In {\em IEEE Conference on Computer Vision and Pattern Recognition
  (CVPR)}, 2017.

\bibitem{breen19983d}
D.~E. Breen, S. Mauch, and R.~T. Whitaker.
\newblock 3d scan conversion of csg models into distance volumes.
\newblock In {\em Proceedings of the 1998 IEEE Symposium on Volume
  Visualization}, pages 7--14, 1998.

\bibitem{chang2015shapenet}
A.~X. Chang, T. Funkhouser, L. Guibas, P. Hanrahan, Q. Huang, Z. Li, S.
  Savarese, M. Savva, S. Song, H. Su, et~al.
\newblock Shapenet: An information-rich 3d model repository.
\newblock {\em arXiv preprint arXiv:1512.03012}, 2015.

\bibitem{chevalier2003segmentation}
L. Chevalier, J. Jaillet, and A. Baskurt.
\newblock Segmentation and superquadric modeling of {3D} objects.
\newblock In {\em The 11-th International Conference in Central Europe on
  Computer Graphics, Visualization and Computer Vision'2003, {WSCG} 2003},
  2003.

\bibitem{chou2022gensdf}
G. Chou, I. Chugunov, and F. Heide.
\newblock Gensdf: Two-stage learning of generalizable signed distance
  functions.
\newblock In {\em Advances in Neural Information Processing Systems (NeurIPS)},
  2022.

\bibitem{choy}
C.~B. Choy, D. Xu, J. Gwak, K. Chen, and S. Savarese.
\newblock 3d-r2n2: A unified approach for single and multi-view 3d object
  reconstruction.
\newblock In {\em Proceedings of the European Conference on Computer Vision
  (ECCV)}, pages 628--644. Springer International Publishing, 2016.

\bibitem{curless1996volumetric}
B. Curless and M. Levoy.
\newblock A volumetric method for building complex models from range images.
\newblock In {\em Proceedings of the 23rd Annual Conference on Computer
  Graphics and Interactive Techniques}, pages 303--312, 1996.

\bibitem{dai2018scancomplete}
A. Dai, D. Ritchie, M. Bokeloh, S. Reed, J. Sturm, and M. Nie{\ss}ner.
\newblock Scancomplete: Large-scale scene completion and semantic segmentation
  for 3d scans.
\newblock In {\em Proceedings of the IEEE Conference on Computer Vision and
  Pattern Recognition (CVPR)}, pages 4578--4587, 2018.

\bibitem{Dai_2017_CVPR}
A. Dai, C. Ruizhongtai~Qi, and M. Niessner.
\newblock Shape completion using 3d-encoder-predictor cnns and shape synthesis.
\newblock In {\em Proceedings of the IEEE Conference on Computer Vision and
  Pattern Recognition (CVPR)}, July 2017.

\bibitem{DEMP1977}
A.~P. Dempster, N.~M. Laird, and D.~B. Rubin.
\newblock Maximum likelihood from incomplete data via the {EM} algorithm.
\newblock {\em Journal of the Royal Statistical Society: Series B
  (Methodological)}, 39(1):1--22, 1977.

\bibitem{cvxnet}
B. Deng, K. Genova, S. Yazdani, S. Bouaziz, G. Hinton, and A. Tagliasacchi.
\newblock Cvxnet: Learnable convex decomposition.
\newblock In {\em 2020 IEEE/CVF Conference on Computer Vision and Pattern
  Recognition (CVPR)}, pages 31--41, 2020.

\bibitem{dong2018psdf}
W. Dong, Q. Wang, X. Wang, and H. Zha.
\newblock Psdf fusion: Probabilistic signed distance function for on-the-fly 3d
  data fusion and scene reconstruction.
\newblock In {\em Proceedings of the European Conference on Computer Vision
  (ECCV)}, pages 701--717, 2018.

\bibitem{shapetemplates}
K. Genova, F. Cole, D. Vlasic, A. Sarna, W. Freeman, and T. Funkhouser.
\newblock Learning shape templates with structured implicit functions.
\newblock In {\em 2019 IEEE/CVF International Conference on Computer Vision
  (ICCV)}, pages 7153--7163, 2019.

\bibitem{GrossICCV88}
A.~D. Gross and T.~E. Boult.
\newblock Error of fit measures for recovering parametric solids.
\newblock In {\em Proceedings of the IEEE/CVF International Conference on
  Computer Vision (ICCV)}, pages 690--694, 1988.

\bibitem{hao2020dualsdf}
Z. Hao, H. Averbuch-Elor, N. Snavely, and S. Belongie.
\newblock Dualsdf: Semantic shape manipulation using a two-level
  representation.
\newblock In {\em Proceedings of the IEEE/CVF Conference on Computer Vision and
  Pattern Recognition (CVPR)}, pages 7631--7641, 2020.

\bibitem{izadi2011kinectfusion}
S. Izadi, D. Kim, O. Hilliges, D. Molyneaux, R. Newcombe, P. Kohli, J. Shotton,
  S. Hodges, D. Freeman, A. Davison, et~al.
\newblock Kinectfusion: real-time 3d reconstruction and interaction using a
  moving depth camera.
\newblock In {\em Proceedings of the 24th annual ACM symposium on User
  interface software and technology}, pages 559--568, 2011.

\bibitem{core_systems_in_human_cognition}
D.~K. Katherine and S.~S. Elizabeth.
\newblock Core systems in human cognition.
\newblock In {\em From Action to Cognition}, volume 164 of {\em Progress in
  Brain Research}, pages 257--264. Elsevier, 2007.

\bibitem{NEURIPS2020_59a3adea}
Y. Kawana, Y. Mukuta, and T. Harada.
\newblock Neural star domain as primitive representation.
\newblock In H. Larochelle, M. Ranzato, R. Hadsell, M.F. Balcan, and H. Lin,
  editors, {\em Advances in Neural Information Processing Systems}, volume~33,
  pages 7875--7886. Curran Associates, Inc., 2020.

\bibitem{leonardis1997superquadrics}
A. Leonardis, A. Jaklic, and F. Solina.
\newblock Superquadrics for segmenting and modeling range data.
\newblock {\em IEEE Transactions on Pattern Analysis and Machine Intelligence},
  19(11):1289--1295, 1997.

\bibitem{liao2018deep}
Y. Liao, S. Donne, and A. Geiger.
\newblock Deep marching cubes: Learning explicit surface representations.
\newblock In {\em Proceedings of the IEEE Conference on Computer Vision and
  Pattern Recognition (CVPR)}, pages 2916--2925, 2018.

\bibitem{Liu_2022_CVPR}
W. Liu, Y. Wu, S. Ruan, and G.~S. Chirikjian.
\newblock Robust and accurate superquadric recovery: A probabilistic approach.
\newblock In {\em Proceedings of the IEEE/CVF Conference on Computer Vision and
  Pattern Recognition (CVPR)}, pages 2676--2685, June 2022.

\bibitem{lorensen1987marching}
W.~E. Lorensen and H.~E. Cline.
\newblock Marching cubes: A high resolution 3d surface construction algorithm.
\newblock {\em ACM siggraph computer graphics}, 21(4):163--169, 1987.

\bibitem{newcombe2015dynamicfusion}
R.~A. Newcombe, D. Fox, and S.~M. Seitz.
\newblock Dynamicfusion: Reconstruction and tracking of non-rigid scenes in
  real-time.
\newblock In {\em Proceedings of the IEEE Conference on Computer Vision and
  Pattern Recognition (CVPR)}, pages 343--352, 2015.

\bibitem{cube_cvpr2018}
C. Niu, J. Li, and K. Xu.
\newblock {Im2Struct}: Recovering {3D} shape structure from a single {RGB}
  image.
\newblock In {\em Proceedings of the IEEE Conference on Computer Vision and
  Pattern Recognition (CVPR)}, June 2018.

\bibitem{park2019deepsdf}
J.~J. Park, P. Florence, J. Straub, R. Newcombe, and S. Lovegrove.
\newblock Deepsdf: Learning continuous signed distance functions for shape
  representation.
\newblock In {\em Proceedings of the IEEE/CVF Conference on Computer Vision and
  Pattern Recognition (CVPR)}, pages 165--174, 2019.

\bibitem{paschalidou2020learning}
D. Paschalidou, L.~V. Gool, and A. Geiger.
\newblock Learning unsupervised hierarchical part decomposition of {3D} objects
  from a single {RGB} image.
\newblock In {\em Proceedings of the IEEE/CVF Conference on Computer Vision and
  Pattern Recognition (CVPR)}, June 2020.

\bibitem{Paschalidou_2021_CVPR}
D. Paschalidou, A. Katharopoulos, A. Geiger, and S. Fidler.
\newblock Neural parts: Learning expressive 3d shape abstractions with
  invertible neural networks.
\newblock In {\em Proceedings of the IEEE/CVF Conference on Computer Vision and
  Pattern Recognition (CVPR)}, pages 3204--3215, June 2021.

\bibitem{paschalidou2019superquadrics}
D. Paschalidou, A.~O. Ulusoy, and A. Geiger.
\newblock Superquadrics revisited: Learning {3D} shape parsing beyond cuboids.
\newblock In {\em Proceedings of the IEEE/CVF Conference on Computer Vision and
  Pattern Recognition (CVPR)}, June 2019.

\bibitem{pentland1987perceptual}
A. Pentland.
\newblock Perceptual organization and the representation of natural form.
\newblock {\em Artificial Intelligence}, 28(3):293--331, 1986.

\bibitem{quispe2015exploiting}
A.~H. Quispe, B. Milville, M.~A. Gutiérrez, C. Erdogan, M. Stilman, H.
  Christensen, and H.~B. Amor.
\newblock Exploiting symmetries and extrusions for grasping household objects.
\newblock In {\em 2015 IEEE International Conference on Robotics and Automation
  (ICRA)}, pages 3702--3708, 2015.

\bibitem{rao2022patchcomplete}
Y. Rao, Y. Nie, and A. Dai.
\newblock Patchcomplete: Learning multi-resolution patch priors for 3d shape
  completion on unseen categories.
\newblock {\em Advances in Neural Information Processing Systems (NeurIPS)},
  2022.

\bibitem{sipu_tro}
S. Ruan, K.~L. Poblete, H. Wu, Q. Ma, and G.~S. Chirikjian.
\newblock Efficient path planning in narrow passages for robots with
  ellipsoidal components.
\newblock {\em IEEE Transactions on Robotics (TRO)}, pages 1--18, 2022.

\bibitem{sipu_ral}
S. Ruan, X. Wang, and G.~S. Chirikjian.
\newblock Collision detection for unions of convex bodies with smooth
  boundaries using closed-form contact space parameterization.
\newblock {\em IEEE Robotics and Automation Letters (RAL)}, 7(4):9485--9492,
  2022.

\bibitem{sharma2022prifit}
G. Sharma, B. Dash, A. RoyChowdhury, M. Gadelha, M. Loizou, L. Cao, R. Wang, E.
  Learned-Miller, S. Maji, and E. Kalogerakis.
\newblock Prifit: Learning to fit primitives improves few shot point cloud
  segmentation.
\newblock In {\em Computer Graphics Forum}, volume~41, pages 39--50. Wiley
  Online Library, 2022.

\bibitem{smirnov2020deep}
D. Smirnov, M. Fisher, V.~G. Kim, R. Zhang, and J. Solomon.
\newblock Deep parametric shape predictions using distance fields.
\newblock In {\em Proceedings of the IEEE/CVF Conference on Computer Vision and
  Pattern Recognition (CVPR)}, pages 561--570, 2020.

\bibitem{solina1990recovery}
F. Solina and R. Bajcsy.
\newblock Recovery of parametric models from range images: the case for
  superquadrics with global deformations.
\newblock {\em IEEE Transactions on Pattern Analysis and Machine Intelligence},
  12(2):131--147, 1990.

\bibitem{sun2019learning}
C. Sun, Q. Zou, X. Tong, and Y. Liu.
\newblock Learning adaptive hierarchical cuboid abstractions of 3d shape
  collections.
\newblock {\em ACM Transactions on Graphics (TOG)}, 38(6):1--13, 2019.

\bibitem{Tulsiani_2017_CVPR}
S. Tulsiani, H. Su, L.~J. Guibas, A.~A. Efros, and J. Malik.
\newblock Learning shape abstractions by assembling volumetric primitives.
\newblock In {\em Proceedings of the IEEE Conference on Computer Vision and
  Pattern Recognition (CVPR)}, July 2017.

\bibitem{superquadrics_num_stable}
N. Vaskevicius and A. Birk.
\newblock Revisiting superquadric fitting: A numerically stable formulation.
\newblock {\em IEEE Transactions on Pattern Analysis and Machine Intelligence},
  41(1):220--233, 2019.

\bibitem{vasutalabot2022hybridsdf}
S. Vasu, N. Talabot, A. Lukoianov, P. Baqu\'e, J. Donier, and P. Fua.
\newblock Hybridsdf: Combining deep implicit shapes and geometric primitives
  for 3d shape representation and manipulation.
\newblock In {\em International Conference on 3D Vision (3DV)}, 2022.

\bibitem{vezzani2017grasping}
G. Vezzani, U. Pattacini, and L. Natale.
\newblock A grasping approach based on superquadric models.
\newblock In {\em 2017 IEEE International Conference on Robotics and Automation
  (ICRA)}, pages 1579--1586, 2017.

\bibitem{vezzani2018improving}
G. Vezzani, U. Pattacini, G. Pasquale, and L. Natale.
\newblock Improving superquadric modeling and grasping with prior on object
  shapes.
\newblock In {\em 2018 IEEE International Conference on Robotics and Automation
  (ICRA)}, pages 6875--6882, 2018.

\bibitem{Weder_2020_CVPR}
S. Weder, J.~L. Sch\"onberger, M. Pollefeys, and M.~R. Oswald.
\newblock Routedfusion: Learning real-time depth map fusion.
\newblock In {\em IEEE/CVF Conference on Computer Vision and Pattern
  Recognition (CVPR)}, June 2020.

\bibitem{wu2022primitive}
Y. Wu, W. Liu, S. Ruan, and G.~S. Chirikjian.
\newblock Primitive-based shape abstraction via nonparametric bayesian
  inference.
\newblock In {\em Proceedings of the European Conference on Computer Vision
  (ECCV)}, pages 479--495. Springer Nature Switzerland, 2022.

\bibitem{yang2021unsupervised}
K. Yang and X. Chen.
\newblock Unsupervised learning for cuboid shape abstraction via joint
  segmentation from point clouds.
\newblock {\em ACM Transactions on Graphics (TOG)}, 40(4):1--11, 2021.

\bibitem{zhou2013elastic}
Q. Zhou, S. Miller, and V. Koltun.
\newblock Elastic fragments for dense scene reconstruction.
\newblock In {\em Proceedings of the IEEE International Conference on Computer
  Vision (ICCV)}, pages 473--480, 2013.

\bibitem{cube_cvpr2017}
C. Zou, E. Yumer, J. Yang, D. Ceylan, and D. Hoiem.
\newblock {3D-PRNN}: Generating shape primitives with recurrent neural
  networks.
\newblock In {\em Proceedings of the IEEE International Conference on Computer
  Vision (ICCV)}, Oct 2017.

\end{thebibliography}
}

\end{document}


\title{Supplementary Material of Marching-Primitives: \\Shape Abstraction from Signed Distance Function}

\author{Weixiao Liu$^{1, 2}$ \quad Yuwei Wu$^{1}$ \quad Sipu Ruan$^{1}$ \quad Gregory S. Chirikjian$^{1}$\\
$^1$National University of Singapore \quad $^2$Johns Hopkins University\\
{\tt\small \{mpewxl, yw.wu, ruansp, mpegre\}@nus.edu.sg}
}

\maketitle

\begin{abstract}
In this supplementary material, we provide the details about the derivations, discussions and experiment settings.
In Sec. \ref{sec:approxsdf}, we provide the detailed derivation of the approximation of the SDF of a superquadric.
In Sec. \ref{sec:initialization}, the primitive initialization strategy in the connectivity marching step is detailed.
In Sec. \ref{sec:derivation}, we provide the derivation of the probabilistic primitive marching step.
Furthermore, Sec. \ref{sec:removal} elaborates the fail-safe primitive removal criterion.
The overview of the Marching-Primitive algorithm is summarized into pseudo-code in Sec. \ref{sec:pseudo-code}.
Finally, in Sec. \ref{sec:implementation}, we detail the experiment implementation and show more qualitative examples.
  
\end{abstract}

\section{Approximation of superquadric SDF}
\label{sec:approxsdf}

\begin{figure} [!ht]
    \centering
    \includegraphics[width=0.8\columnwidth]{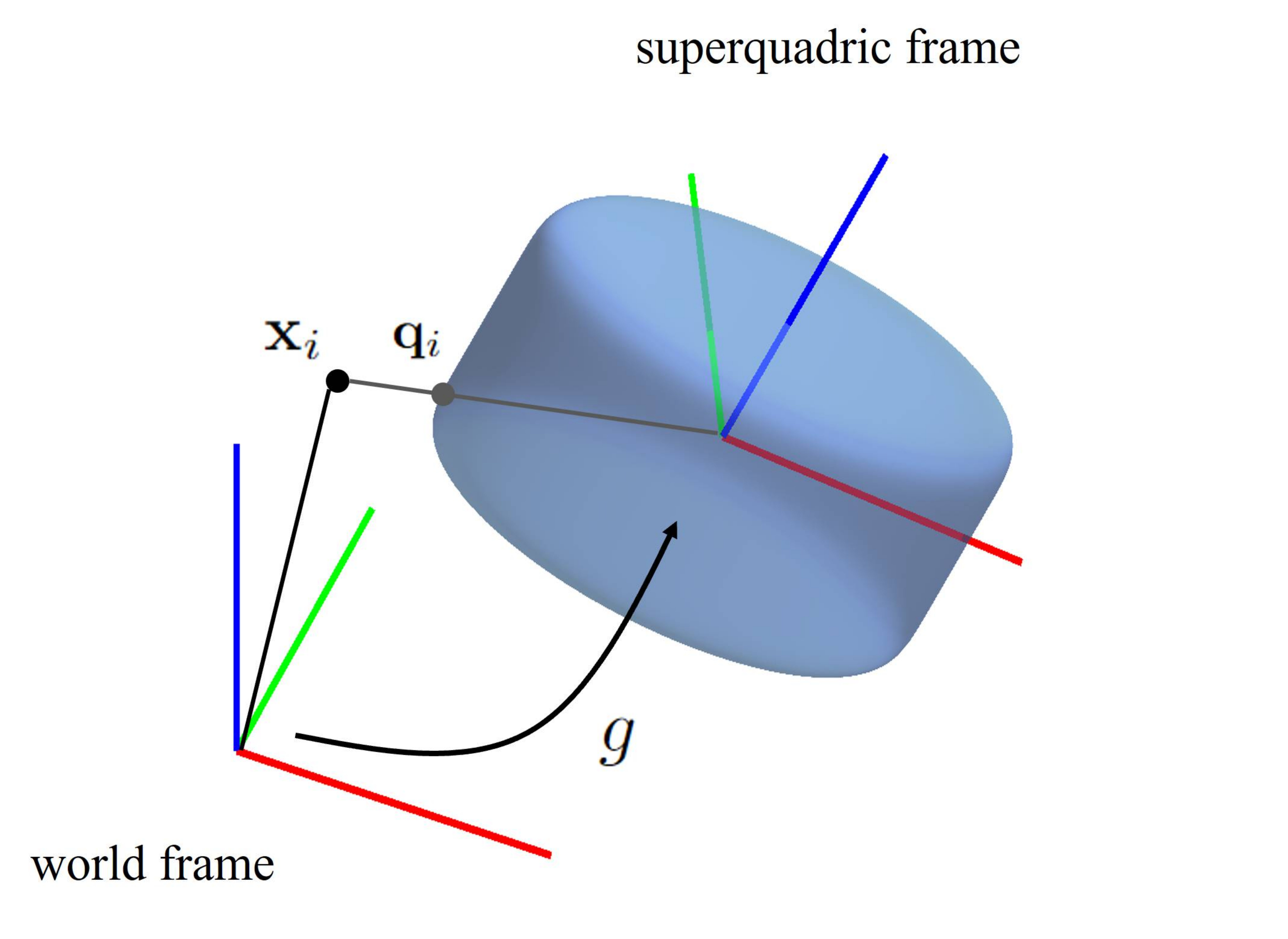} 
    \caption{Signed distance approximated by signed radial distance.}
    \label{fig:approxSDF}
    \vspace{-0.5cm}
\end{figure}

Recall the signed radial distance of a point $\mathbf{x}_i$ to a general-posed superquadric in Eq. (2) of the paper
\begin{equation}
d_{\boldsymbol{\theta}}(\mathbf{x}_i)=\left( 1 - f^{-\frac{\epsilon_1}{2}}(g^{-1}\circ\mathbf{x}_i) \right)\|g^{-1}\circ\mathbf{x}_i\|_{2}
\label{eqn:radial_distance_re}
\end{equation}
The derivation is as follows (with an illustration shown in Fig.\ref{fig:approxSDF}).
The radial distance of a point $\mathbf{x}_i$ to the surface of a superquadric is defined as 
\begin{equation}
    \|\mathbf{x}_i-\mathbf{q}_i\|_2
\end{equation}
$\mathbf{q}_i$ is where the vector from the center of the superquadric frame to $\mathbf{x}_i$ intersects the surface.
Therefore, when viewed from the superquadric frame $g$, the vectors $g^{-1}\circ \mathbf{q}_i$ and $g^{-1}\circ \mathbf{x}_i$ are colinear, \ie 
\begin{equation}
    g^{-1}\circ \mathbf{q}_i = \alpha (g^{-1}\circ \mathbf{x}_i)\text{, where }\alpha\in \mathbb{R}
\end{equation}
Note that $g^{-1}\circ \mathbf{q}_i$ lies on the surface of superquadric.
Thus, substituting it into the implicit equation of the superquadric (Eq. (1) in the paper), we obtain 
\begin{equation}
    \alpha = f^{-\frac{\epsilon_1}{2}}(g^{-1}\circ\mathbf{x}_i)
\end{equation}
Therefore,
\begin{equation}
\begin{aligned}
    \|\mathbf{x}_i-\mathbf{q}_i\|_2=&\|g^{-1}\circ\mathbf{x}_i-g^{-1}\circ\mathbf{q}_i\|_2\\
    =&\|(1-\alpha)g^{-1}\circ\mathbf{x}_i\|_2
\end{aligned}
\end{equation}
Considering the inside/outside of $\mathbf{x}_i$ relative to the superquadric surface, the signed radial distance is thus Eq.\eqref{eqn:radial_distance_re}.

\section{Primitive Initialization}
\label{sec:initialization}
\begin{figure} [!tp]
    \centering
    \includegraphics[width=0.65\columnwidth]{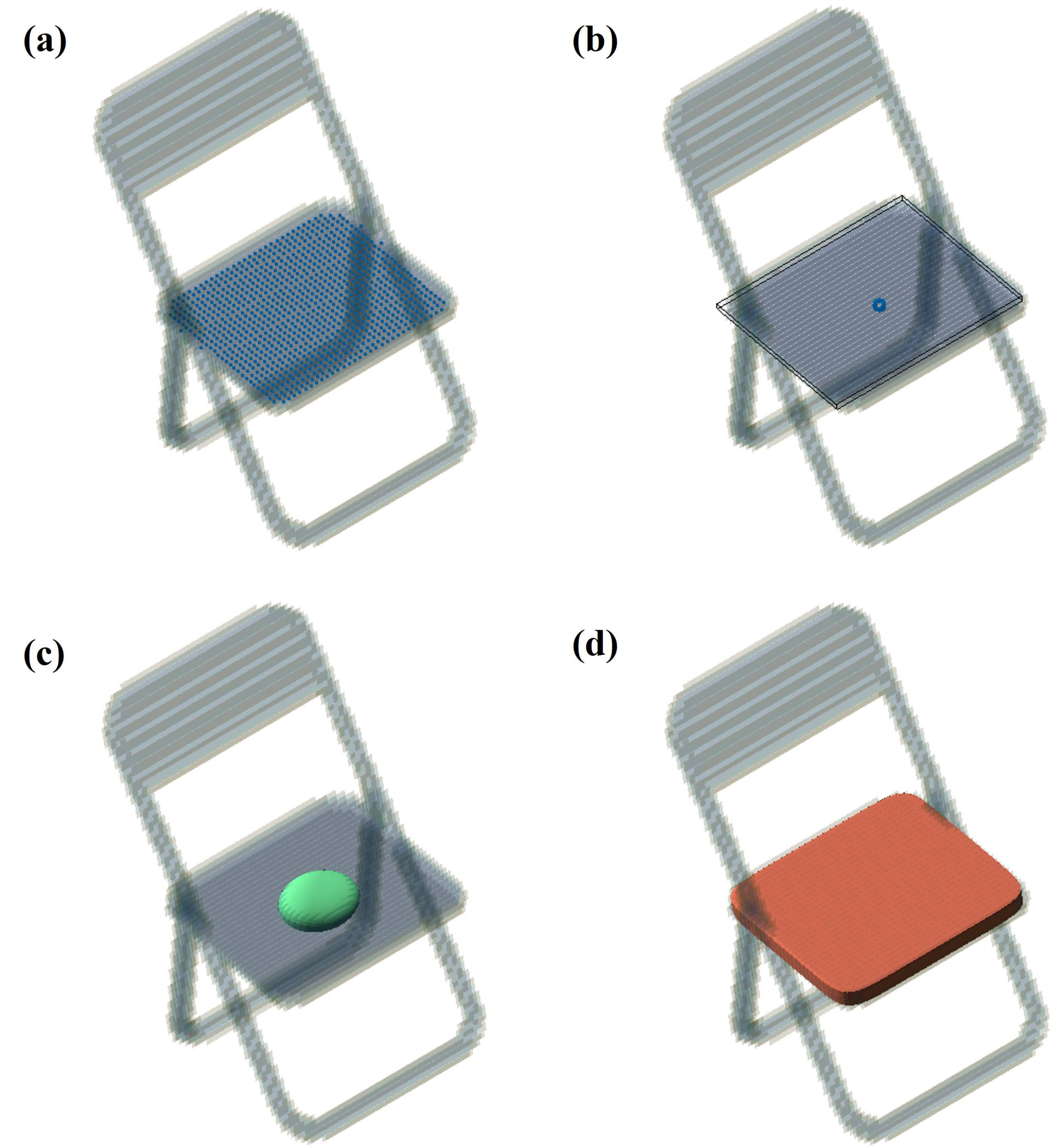} 
    \caption{Visualizations of concepts. (a) Blue dots indicate the detected VOI. (b) The smallest bounding-box encompassing the VOI. (c) Initialization of the primitive as an ellipsoid. (d) The final marched superquadric capturing the local geometry.}
    \label{fig:conceptsupp}
    \vspace{-0.5cm}
\end{figure}

In Sec. 3.1 of the paper, the geometric primitive is initialized as an ellipsoid for each volume of interest (VOIs) detected in the target SDF.
This is achieved by first finding out the smallest bounding-box encompassing the connected voxels that form each of the VOI.
For example, the lengths of a bounding-box are $l_x$, $l_x$ and $l_z$, and the centroid locates at $\mathbf{x}_c$.
Ideally, the primitive is initialized as an ellipsoid centered at $\mathbf{x}_c$ with scales proportional to the lengths correspondingly.
Recall that a superquadric $\boldsymbol{\theta}$ is parameterized by
\begin{equation}
    \boldsymbol{\theta}= \{\epsilon_1, \epsilon_2, a_x, a_y, a_z, \mathbf{R}, \mathbf{t}\}
\end{equation}
Then, the initialized ellipsoid is a special case of the superquadrics
\begin{equation}
    \boldsymbol{\theta}_{init}=\{1, 1, \gamma l_x, \gamma l_y, \gamma l_z, \mathbf{I}, \mathbf{x}_c\}
\end{equation}
where $\gamma$ is the initial scale ratio, $\mathbf{I}$ is the identity rotation matrix.
However, if the VOI is nonconvex, the centroid might lie in the exterior space.
In this situation, it has the risk of activating the auto-degeneration mechanism.
Therefore, instead of $\mathbf{t}=\mathbf{x}_c$, we constrain the initial location $\mathbf{t}$ within the interior of the target shape:
\begin{equation}
    \mathbf{t}=\argmin_{\mathbf{x}_i\in\mathbf{V},d(\mathbf{x}_i)\leq 0}\|\mathbf{x}_i-\mathbf{x}_c\|_2
\end{equation}
where $\mathbf{V}$ is the current set of voxel points, $d(\mathbf{x}_i)$ is the target signed distance evaluated at voxel point $\mathbf{x}_i$.
$\boldsymbol{\theta}_{init}$ works as the initial input to the subsequent probabilistic primitive marching step.
The concepts are visualized in Fig.\ref{fig:conceptsupp} for better understanding.

\section{Derivation of the Probabilistic Marching}
\label{sec:derivation}
In this section, we provide the detailed derivation of the probabilistic primitive marching in Sec. 3.4 of the paper.
Based on the probabilistic model $p\big(d(\mathbf{x}_i)|\boldsymbol{\theta}_k, z_{ik}\big)$ (Eq.8 in the paper), the likelihood of the superquadric parameter $\boldsymbol{\theta}_k$ and variance $\sigma^2$ given the target SDF is
\begin{equation}
\begin{split}
    L(\boldsymbol{\theta}_k, \sigma^2)=&\prod_{\mathbf{x}_i\in\mathbf{V}}p(d(\mathbf{x}_i), z_{ik}|\boldsymbol{\theta}_k)\\
    =&\prod_{\mathbf{x}_i\in\mathbf{V}}p\big(d(\mathbf{x}_i)|\boldsymbol{\theta}_k, z_{ik}\big)p(z_{ik}|\boldsymbol{\theta}_k)\\
    =&\prod_{\mathbf{x}_i\in\mathbf{V}}p_0(\mathbf{x}_i)^{1-z_{ik}}\mathcal{N}\big(d_i|d_{\boldsymbol{\theta}_k}(\mathbf{x}_i), \sigma^2\big)^{z_{ik}}p(z_{ik})
\end{split}
\end{equation}
where
\begin{equation}
    p_0(\mathbf{x}_i) = \frac{\mathds{1}_{d_i\in[-t,0)}}{t} \quad d_i\doteq d(\mathbf{x}_i)
\end{equation}
and $p(z_{ik})$ is the prior probability of the correspondence between the $i$th voxel point and the $k$th primitive, which is independent of $\boldsymbol{\theta}_k$, \ie $p(z_{ik}|\boldsymbol{\theta}_k)=p(z_{ik})$.
As discussed in the paper, we assume that $z_{ik}$ is subjected to a Bernoulli prior distribution $B(p_0)$, \ie $p(z_{ik}=1)=p_0$.
The definitions of other variables can be found in the paper.
Our goal is to find the optimal $\boldsymbol{\theta}_k$ and $\sigma^2$ that maximize the likelihood function.
This is equivalent to minimizing the negative log-likelihood
\begin{equation}
\begin{split}
    l(\boldsymbol{\theta}_k, \sigma^2)=&-\log L(\boldsymbol{\theta}_k, \sigma^2)\\
    =&-\sum_{\mathbf{x}_i\in \mathbf{V}}\log \Bigg[p(z_{ik})p_0(\mathbf{x}_i)^{1-z_{ik}}c(\sigma^2)^{z_{ik}}\\
    &\exp\bigg(-\frac{1}{2}\Big(\frac{d(\mathbf{x}_i)-d_{\boldsymbol{\theta}_k}(\mathbf{x}_i)}{\sigma}\Big)^2 \bigg)^{z_{ik}}\Bigg]\\
    =&\sum_{\mathbf{x}_i\in \mathbf{V}} \bigg[\frac{z_{ik}}{2}\Big(\frac{d(\mathbf{x}_i)-d_{\boldsymbol{\theta}_k}(\mathbf{x}_i)}{\sigma}\Big)^2-\log p(z_{ik})\\
    &-z_{ik}\log c(\sigma^2)-(1-z_{ik})\log p_0(\mathbf{x}_i)\bigg]
\end{split}
\end{equation}
where $c(\sigma^2)=(2\pi \sigma^2)^{-\frac{1}{2}}$ is the normalizing coefficient of the Gaussian distribution.
By ignoring the terms independent of $\boldsymbol{\theta}_k$ and $\sigma^2$, it is equivalent to minimize
\begin{equation}
    l'(\boldsymbol{\theta}_k, \sigma^2)=\sum_{\mathbf{x}_i\in \mathbf{V}} z_{ik}\bigg[\frac{\big(d(\mathbf{x}_i)-d_{\boldsymbol{\theta}_k}(\mathbf{x}_i)\big)^2}{2\sigma^2}-\log c(\sigma^2)\bigg]
\label{eqn:mle}
\end{equation}
Unlike $d(\mathbf{x}_i)$ which is observed from the target signed distance, the correspondence $z_{ik}$ is a latent variable that cannot be observed.
Therefore, it is intractable to solve Eq.\ref{eqn:mle} directly.
Our algorithm solves the problem in a two-step expectation-maximization fashion.
That is, $z_{ik}$ is replaced with 
\begin{equation}
    P_{ik}\doteq E\big(z_{ik}|\boldsymbol{\theta}_k, d(\mathbf{x}_i)\big)=p\big(z_{ik}=1|\boldsymbol{\theta}_k, d(\mathbf{x}_i)\big)
\label{eqn:expect}
\end{equation}
Eq.\ref{eqn:expect} is the conditional expectation of $z_{ik}$ given the current estimation of $\boldsymbol{\theta}_k$ and the target SDF, whose value can be calculated by Eq.9 in the paper.
Subsequently, we derive that the minimization of Eq.\ref{eqn:mle} is equivalent to Eq.10 in the paper, where we use an adaptive activation subset $\mathbf{V}_a$ instead of the whole voxel space $\mathbf{V}$ to boost performance.
After we obtain the updated primitive estimation $\boldsymbol{\theta}_k$, the variance $\sigma^2$ of the Gaussian distribution can be updated in closed form by solving
\begin{equation}
\begin{split}
    \quad&\frac{\partial l'}{\partial \sigma^2}=0\\
    \Leftrightarrow
    &\sum_{\mathbf{x}_i\in \mathbf{V}_a}P_{ik}
    \bigg[\frac{\big(d(\mathbf{x}_i)-d_{\boldsymbol{\theta}_k}(\mathbf{x}_i)\big)^2-\sigma^2}{2\sigma^4}\bigg]=0\\
    \Leftrightarrow
    &\quad \sigma^2=\frac{\sum_{\mathbf{x}_i\in \mathbf{V}_a}P_{ik}\big(d(\mathbf{x}_i)-d_{\boldsymbol{\theta}_k}(\mathbf{x}_i)\big)^2}{\sum_{{\mathbf{x}_i\in \mathbf{V}_a}}P_{ik}}
\end{split}
\end{equation}

\section{Primitive Removal Criterion}
\label{sec:removal}
In this section, we detail the fail-safe primitive removal criterion introduced in Sec. 3.5 in the paper.
Our method counts the number of positive (exterior), negative (interior), and inactive voxels encompassed by the recovered primitive, which we denote as $N_+$, $N_-$ and $N_0$, respectively.
The inactive voxels are those already fitted by recovered primitives, which is defined by Eq.7 in the paper.
Our algorithm removes the primitive from the representation if
\begin{equation}
    N_-<1 \quad or\quad \frac{N_+}{N_+ +N_- + N_0}\geq 0.5
\end{equation}
The first criterion removes the auto-degenerated primitive that shrinks to a point.
The second one is a fail-safe checking criterion, which removes the primitive that significantly contradicts the target SDF.

\section{Overview of the Algorithm}
\label{sec:pseudo-code}
In this section, we briefly summarize the Marching-Primitives algorithm into a pseudo-code (Algorithm.\ref{code:algorithm}) to give an overview of the structure.
Note that $V_-$ in the fourth row indicates the sets of voxels with negative signed distance, \ie interior of the shape.
The Marching-Primitives can be roughly separated into 2 parts.
Firstly, it marches on the signed distance domain (row 5-15) to find VOIs by analysing the connectivity.
Then for each VOI, the algorithm continues to march on the voxelized space domain (row 16-24) to grow a primitive capturing the local geometry of the VOI.
The algorithm terminates when all the interior volumes are well captured by the primitive representation $\boldsymbol{\Theta}$.

\begin{algorithm}
\caption{Marching-Primitives}\label{alg:marching}
\begin{algorithmic}[1]
    \State \textbf{Input:} voxel set $\mathbf{V}$, with target SDF $d(\cdot)$
    \State \textbf{Output:} primitive set $\boldsymbol{\Theta}$
    \State $\boldsymbol{\Theta} \gets \{\}$
    \While{$\mathbf{V}_-\neq\emptyset$}
         \State generate marching sequence $T^c$ \Comment{Eq.4 in paper}
         \For{$t_m^c$ in $T^c$}
             \If{$t_m^c>$ termination threshold}
                 \State \textbf{return} $\boldsymbol{\Theta}$
             \Else
                 \State calculate VOIs $\bar{S}_m$ \Comment{Eq.6 in paper}
                 \If{$\bar{S}_m\neq \emptyset$}
                    \State\textbf{break for}
                 \EndIf
             \EndIf
         \EndFor
         \For{$\mathcal{S}_k$ in $\bar{S}_m$}
             \State initialize primitive $\boldsymbol{\theta}_k^{init}$ \Comment{Eq.7 in supplement}
             \While{not converged}
                 \State march correspondence $P_{ik}$ \Comment{Eq.9 in paper}
                 \State update primitive $\boldsymbol{\theta}_k$ \Comment{Eq.10 in paper}
             \EndWhile
             \State $\boldsymbol{\theta}_k \rightarrow \boldsymbol{\Theta}$ \textbf{if} $\boldsymbol{\theta}_k$ valid \Comment{Eq.15 in supplement}
             \State $\mathbf{V}=\mathbf{V}-\{\mathbf{x}_i, d(\mathbf{x}_i)\leq 0 \land d_{\boldsymbol{\theta}_k}(\mathbf{x}_i)\leq0\}$
         \EndFor
    \EndWhile
    \State \textbf{return} $\boldsymbol{\Theta}$
\end{algorithmic}
\label{code:algorithm}
\end{algorithm}

\section{Implementation and Additional Results}
\label{sec:implementation}
\subsection{Metrics}
In this section, we provide details on the two metrics used to evaluate the experiments.

\textbf{Chamfer $L_1\text{-distance}$}:
The common Chamfer $L_1\text{-distance}$ is defined as follows:
\begin{equation}
\label{eq:chamfer}
\begin{split}
 D_{\text {chamfer }}(\mathbf{X}, \mathbf{Y})  = & \frac{1}{M} \sum_{\mathbf{y}_{j} \in \mathbf{Y}} \min _{\mathbf{x}_{i} \in \mathbf{X}}\left\|\mathbf{y}_{j}-\mathbf{x}_{i}\right\|_1 +\\
 & \frac{1}{N} \sum_{\mathbf{x}_{i} \in \mathbf{X}} \min _{\mathbf{y}_{j} \in \mathbf{Y}}\left\|\mathbf{x}_{i}-\mathbf{y}_{j}\right\|_1
\end{split}
\end{equation}
where $\mathbf{X}=\{\mathbf{x_i}\}$ denotes the points sampled from the predicted model, $\mathbf{Y}=\{\mathbf{y_j}\}$ denotes the points sampled from the original model, and $N$ and $M$ is the number of points of the sets $\mathbf{X}$ and $\mathbf{Y}$, respectively. 
For D-FAUST dataset, it provides a dense point cloud for each human model, which we take as $\mathbf{Y}$.
ShapeNet, on the other hand, does not provide point cloud representation for the object. 
So, we need to sample points densely on each face of the original mesh.
To obtain $\mathbf{X}$, we apply the equal-distance sampling strategy  \cite{Liu_2022_CVPR} on each superquadric surface $\boldsymbol{\theta}_k \in \boldsymbol{\Theta}$ of the predicted model to get a point set $\boldsymbol{\Gamma}_k$. However, some points from $\boldsymbol{\Gamma}_k$
might lie inside of another superquadric $\boldsymbol{\theta}_l, l \neq k$, \ie, those points are on the inside of the 3D model. Therefore, we need to remove those interior points by forming a subset $\Tilde{\boldsymbol{\Gamma}}_k \subset \boldsymbol{\Gamma}_k$,
\begin{equation}
    \Tilde{\boldsymbol{\Gamma}}_k =\{ \boldsymbol{\gamma}^k_i \mid \boldsymbol{\gamma}^k_i \in \boldsymbol{\Gamma}_k, f(\boldsymbol{\gamma}^k_i, \boldsymbol{\theta}_l) \geq 0, \forall \boldsymbol{\theta}_l \in \boldsymbol{\Theta} \}
\end{equation}
where $f(.)$ denotes the inside-outside function of the superquadric.
By taking the union of all the point sets $\{\Tilde{\boldsymbol{\Gamma}}_1,\Tilde{\boldsymbol{\Gamma}}_2,...,\Tilde{\boldsymbol{\Gamma}}_K \}$, we obtain a point cloud representation for the predicted model, which we treat as $\mathbf{X}$.
For both ShapeNet and D-FAUST, we further downsample $\mathbf{X}$ and $\mathbf{Y}$ to 50K-60K points for calculating the Chamfer distance.
The first term of Eq.\ref{eq:chamfer} computes how far on average the closest point of the predicted model is to the original mesh, and the second term calculates how far on average the closest point of the original mesh is to the predicted model. 
Thus, a lower value of Chamfer distance implies a better abstraction accuracy in terms of surface fitness. 

\textbf{Intersection over Union (IoU)}:
The definition of IoU is shown as follows:
\begin{equation}
\label{eq:iou}
\operatorname{IoU} = \frac{V\left(S_{\text {pred}} \cap S_{\text {original}}\right)}{V\left(S_{\text {pred}} \cup S_{\text {original}}\right)},
\end{equation}
where $S_{\text {pred}}$ is the predicted primitive-based model obtained by our algorithm, $S_{\text {original}}$ is the original mesh model, and $V(.)$ computes the volume.
It is difficult, if not impossible, to obtain the volume of the intersection or union of two models.
Therefore, we approximate the volume with the Monte Carlo method.
Firstly, we sample a set of points $\boldsymbol{\Phi}$ uniformly with a predefined density inside the bounding box of $S_{\text {pred}} \cup S_{\text {original}}$.
We use $100^3$ points for ShapeNet and $64^3$ points for DFAUST.
The number is far more than the previous papers, expecting a more accurate evaluation.
Then, for each point $\mathbf{x} \in \boldsymbol{\Phi}$, we check if it is inside of the original mesh and the predicted model, respectively.
We approximate  $V\left(S_{\text {pred}} \cap S_{\text {original}}\right)$ to be the number of points that are on the inside of both $S_{\text {original}}$ and $S_{\text {pred}}$, and approximate $V\left(S_{\text {pred}} \cup S_{\text {original}}\right)$ with the number of points that are on the inside of either $S_{\text {original}}$ or $S_{\text {pred}}$. 
If two models match perfectly, the IoU will be 1 and if two models disjoint from each other, the IoU is 0. 

\subsection{Implementation Details}
In this section, we elaborate on the parameters implemented in the experiment.
All the experiments use the settings provided as follows, if not specified in the experiment section in the paper.
The truncation threshold for the target and source SDF is 1.3 times the input grid interval.
In the connectivity marching step, the common ratio of the geometric sequence $\alpha$ is $4/5$;
The minimum size of the valid connected volume $N_c=5$;
The primitive initial scale ratio $\gamma=0.1$;
The terminating marching threshold is 0.01 times the negative truncation threshold.
For the probabilistic model, the parameter of the Bernoulli prior distribution is set as $p_0=0.01$;
The variance $\sigma^2$ is initialized as the truncation threshold.
During the primitive update step, we set the activation distance $a$ as 3.5 times the truncation threshold.
The source code of our algorithm is implemented in MATLAB.
The experiments are conducted on a computer running Intel Core i9-9900K CPU.
The baseline method SQ\cite{paschalidou2019superquadrics} is trained and tested on an NVIDIA RTX3090 GPU.

All the methods consume different types of input.
We use the official codes and configurations of \cite{paschalidou2019superquadrics, wu2022primitive}.
For SQs, occupancy grids of resolution $32^3$ are generated from meshes by the provided code.
We had to and tried to modify their network to consume occupancy grids of $128^3$.
Relatively incremental improvement is observed (\eg, chair IoU $0.30\rightarrow 0.34$).
Therefore, we used the original network and followed the official configuration for consistency with the previous literature.
We use 1000 and 200 points from objects and each superquadric for the loss function, respectively.
For NB, we densely sample points from mesh and uniformly downsample to around 3500 (ShapeNet) and 5500 (DFAUST) points and set the number of initial components $K=30$ following the settings in \cite{wu2022primitive}.

\subsection{Number of Parts}
\begin{figure}[!tp]
  \centering
  \includegraphics[width=1\columnwidth]{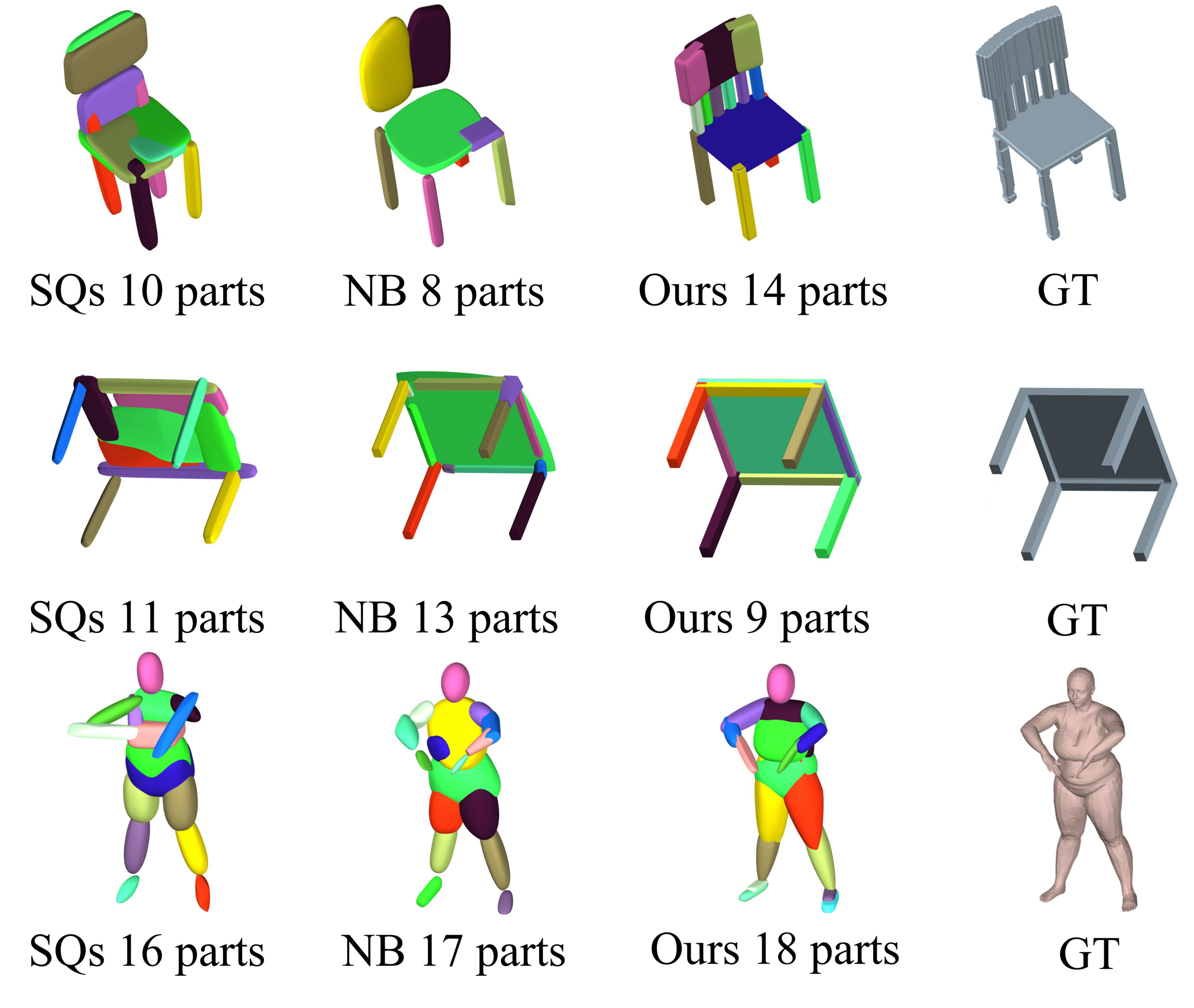} 
  \caption{Shape abstraction results with the number of parts. Recovered primitives are colored in different colors.}
  \label{fig:rebuttal}
  \vspace{-0.5cm}
\end{figure}

The number of parts used is not and cannot be predefined in all the methods.
SQs \cite{paschalidou2019superquadrics} has a hyper-parameter to limit the maximum number set to 20.
The training is unsupervised, and the network learns to predict the number of components.
NB \cite{wu2022primitive} infers the number via the Chinese Restaurant Process, splitting/merging when probabilistically necessary with no limits.
Our method grows parts as needed.
Since there is no ground truth and shapes vary greatly even in the same category, it is hard to quantify the correct number, which makes statistics less meaningful.
Our result is satisfying qualitatively as shown in Fig.\ref{fig:rebuttal}.
Our method can successfully separate different parts if they possess different geometric semantics (\eg telling apart cuboids, cylinders, and balls).
Therefore, it is semantically interpretable.
In many cases (\eg, Fig.\ref{fig:rebuttal}), the segmentation coincides with the human-defined semantics, though not trained to.

\subsection{Time Performance}
The proposed method (MPS) has an average runtime of 6.7s on ShapeNet and 2.5s on DFAUST per item.
Time varies on the complexity of objects and grid resolutions.
For an intuitive example, the chair, table, and human in Fig.\ref{fig:rebuttal} take 3.9s, 3.6s, and 2.8s, respectively.
The complex Reading Room takes 146s for resolution $400^3$ and 30s for $200^3$.

\begin{figure*}[!ht]
    \centering 
    \includegraphics[scale=0.075]{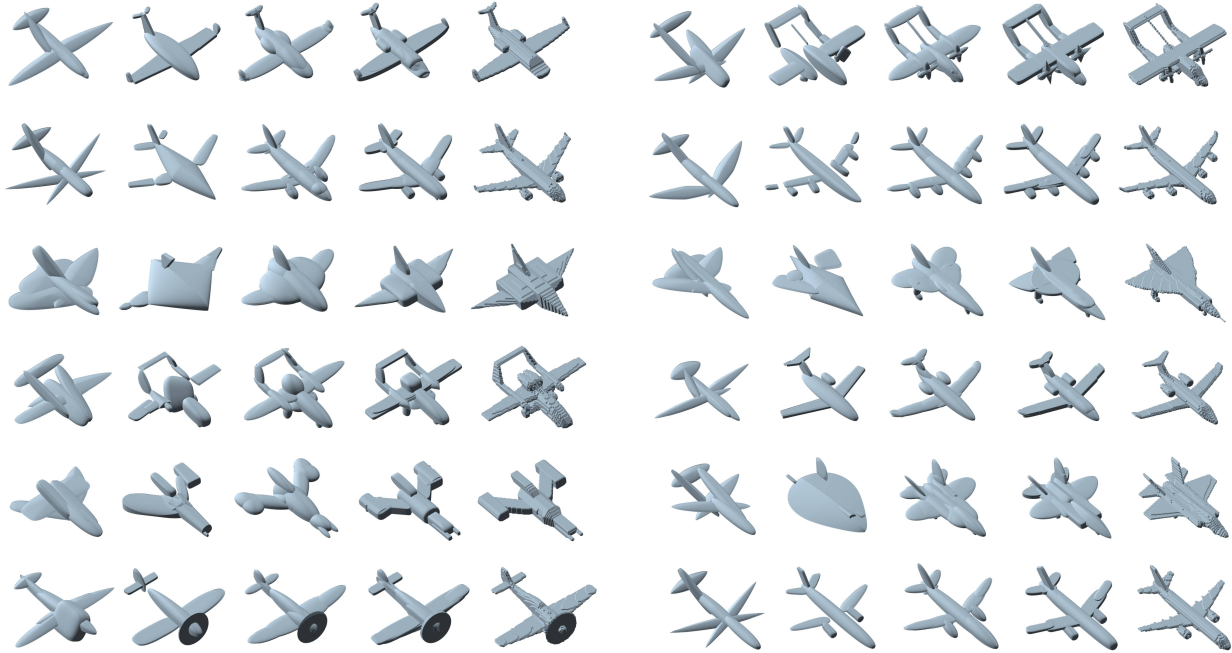} 
    \caption{Shape Abstraction results on airplanes. From left to right: SQs, Non-parametric Bayesian (NB), Marching-Primitives with ellipsoids (MPE), Marching-Primitives with superquadrics (MPS), and the ground truth (pre-processed watertight mesh).}
\vspace{-0.0cm}
\end{figure*}
\begin{figure*}[!ht]
    \centering 
    \includegraphics[scale=0.075]{figures/suppchair_compressed.pdf} 
    \caption{Shape Abstraction results on chairs. From left to right: SQs, Non-parametric Bayesian (NB), Marching-Primitives with ellipsoids (MPE), Marching-Primitives with superquadrics (MPS), and the ground truth (pre-processed watertight mesh).}
\vspace{-0.0cm}
\end{figure*}
\begin{figure*}[!ht]
    \centering 
    \includegraphics[scale=0.078]{figures/suppbenchsofa_compressed.pdf} 
    \caption{Shape Abstraction results on benches and sofas. From left to right: SQs, Non-parametric Bayesian (NB), Marching-Primitives with ellipsoids (MPE), Marching-Primitives with superquadrics (MPS), and the ground truth (pre-processed watertight mesh).}
\vspace{0.5cm}
\end{figure*}
\begin{figure*}[!ht]
    \centering 
    \includegraphics[scale=0.078]{figures/supptable_compressed.pdf} 
    \caption{Shape Abstraction results on tables. From left to right: SQs, Non-parametric Bayesian (NB), Marching-Primitives with ellipsoids (MPE), Marching-Primitives with superquadrics (MPS), and the ground truth (pre-processed watertight mesh).}
\vspace{0.5cm}
\end{figure*}
\begin{figure*}[!ht]
    \centering 
    \includegraphics[scale=0.078]{figures/supplamp_compressed.pdf} 
    \caption{Shape Abstraction results on lamps. From left to right: SQs, Non-parametric Bayesian (NB), Marching-Primitives with ellipsoids (MPE), Marching-Primitives with superquadrics (MPS), and the ground truth (pre-processed watertight mesh).}
\vspace{0.5cm}
\end{figure*}
\begin{figure*}[!ht]
    \centering 
    \includegraphics[scale=0.078]{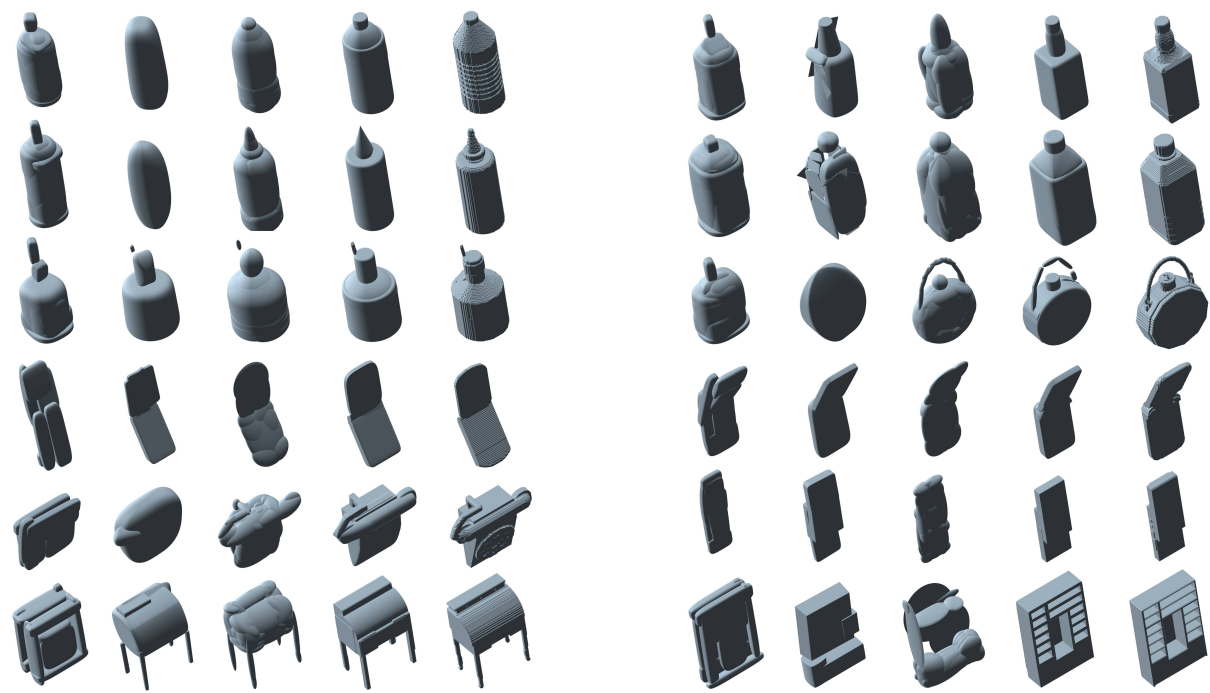} 
    \caption{Shape Abstraction results on bottles, phones and cabinets. From left to right: SQs, Non-parametric Bayesian (NB), Marching-Primitives with ellipsoids (MPE), Marching-Primitives with superquadrics (MPS), and the ground truth.}
\vspace{0.5cm}
\end{figure*}

\begin{figure*}[!ht]
    \centering 
    \includegraphics[scale=0.078]{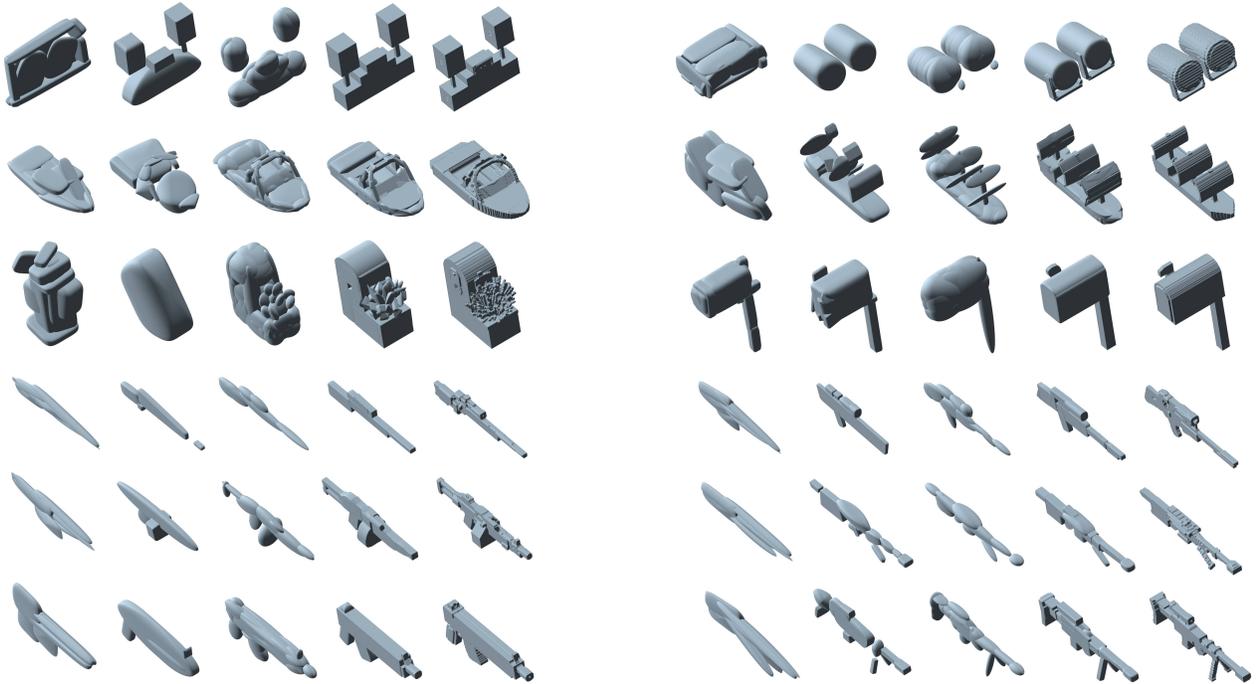} 
    \caption{Shape Abstraction results on speakers, water-crafts, mailboxes and rifles. From left to right: SQs, Non-parametric Bayesian (NB), Marching-Primitives with ellipsoids (MPE), Marching-Primitives with superquadrics (MPS), and the ground truth.}
\vspace{0.5cm}
\end{figure*}

\subsection{Additional Results}
Due to the limited length of the paper, in this Supplementation Material, we prepare more qualitative comparisons on the ShapeNet dataset.
From the additional results, we further demonstrate that our method is able to achieve high accuracy shape abstraction.
Our method not only well captures the geometry of different objects in a same category, but also is generalizable among various categories without the need of fine-tuning.

{\small
\bibliographystyle{ieee_fullname}
\bibliography{egbib}
}